\documentclass[lettersize,journal]{IEEEtran}
\usepackage{amsmath,amsfonts}
\usepackage{algorithmic}
\usepackage{algorithm}
\usepackage{array}
\usepackage[caption=false,font=normalsize,labelfont=sf,textfont=sf]{subfig}
\usepackage{textcomp}
\usepackage{stfloats}
\usepackage{url}
\usepackage{verbatim}
\usepackage{graphicx}
\usepackage{cite}

\usepackage{booktabs}
\usepackage{amssymb}
\usepackage{multirow}
\usepackage{hyperref}   

\usepackage{color}
\usepackage{enumerate}
\newtheorem{definition}{Definition}
\newtheorem{proposition}{Proposition}

\newtheorem{theorem}{Theorem}
\newtheorem{remark}{Remark}

\begin{document}

\title{Improving Generalization in Meta-Learning via Meta-Gradient Augmentation}

\author{Ren Wang, Haoliang Sun$^\ast$, Qi Wei, Xiushan Nie, Yuling Ma, Yilong Yin$^\ast$
\thanks{Ren Wang, Haoliang Sun, Qi Wei, Yilong Yin are with the School of Software, Shandong University, Jinan, China. E-mail: $\{$xxlifelover, hlsun.cn, 1998v7$\}$@gmail.com, ylyin@sdu.edu.cn.}
\thanks{Xiushan Nie, Yuling Ma are with the School of Computer Science and Technology, Shandong Jianzhu University, Jinan, China. E-mail: $\{$niexiushan19, mayuling20$\}$@sdjzu.edu.cn.}
\thanks{$^\ast$Corresponding authors.}}



\markboth{Journal of \LaTeX\ Class Files,~Vol.~14, No.~8, August~2021}%
{Shell \MakeLowercase{\textit{et al.}}: A Sample Article Using IEEEtran.cls for IEEE Journals}

\IEEEpubid{0000--0000/00\$00.00~\copyright~2021 IEEE}

\maketitle

\begin{abstract}
Meta-learning methods typically follow a two-loop framework, where each loop potentially suffers from notorious overfitting, hindering rapid adaptation and generalization to new tasks. Existing schemes solve it by enhancing the mutual-exclusivity or diversity of training samples, but these data manipulation strategies are data-dependent and insufficiently flexible. This work alleviates overfitting in meta-learning from the perspective of gradient regularization and proposes a data-independent \textbf{M}eta-\textbf{G}radient \textbf{Aug}mentation (\textbf{MGAug}) method. The key idea is to first break the rote memories by network pruning to address memorization overfitting in the inner loop, and then the gradients of pruned sub-networks naturally form the high-quality augmentation of the meta-gradient to alleviate learner overfitting in the outer loop. Specifically, we explore three pruning strategies, including \textit{random width pruning}, \textit{random parameter pruning}, and a newly proposed \textit{catfish pruning} that measures a Meta-Memorization Carrying Amount (MMCA) score for each parameter and prunes high-score ones to break rote memories as much as possible. The proposed MGAug is theoretically guaranteed by the generalization bound from the PAC-Bayes framework. In addition, we extend a lightweight version, called MGAug-MaxUp, as a trade-off between performance gains and resource overhead. Extensive experiments on multiple few-shot learning benchmarks validate MGAug's effectiveness and significant improvement over various meta-baselines. The code is publicly available at \url{https://github.com/xxLifeLover/Meta-Gradient-Augmentation}.
\end{abstract}







\section{Introduction}
\IEEEPARstart{M}{eta-learning} aims to rapidly adapt to unseen new tasks by observing the learning process over a wide range of tasks, and has been applied to various scenarios\cite{DBLP:journals/pami/Hospedales22,DBLP:journals/air/HuismanRP21}, including few-shot learning ~\cite{DBLP:journals/tnn/TianLG22,DBLP:journals/pami/SunLCCS22,DBLP:journals/pami/CoskunZTBNTS23}, continual learning ~\cite{DBLP:conf/nips/JavedW19,DBLP:conf/nips/GuptaYP20}, transfer learning ~\cite{DBLP:conf/acl/QianY19,DBLP:conf/aaai/LinW0H20}, etc. Most prevalent meta-learning methods follow a unified two-loop framework ~\cite{DBLP:conf/nips/GoldblumFG20,DBLP:conf/icml/GoldblumRFNCG20}. In the outer loop, a meta-learner explores meta-knowledge from numerous tasks. Based on this meta-knowledge, base learners in the inner loop are expected to quickly fine-tune and adapt to new tasks.

As indicated in ~\cite{DBLP:journals/corr/abs-1907-07287,DBLP:conf/icml/YaoHZ0TZHL21}, the two-loop framework may potentially suffer from meta-overfitting in two aspects: \textit{memorization overfitting} and \textit{learner overfitting}. Memorization overfitting \cite{DBLP:conf/nips/RajendranIJ20} means that the base learner handles tasks merely based on meta-knowledge rather than task-specific fine-tuning in the inner loop. In this way, meta-knowledge degenerates into rote memorization, which hinders rapid adaptation to new tasks. Learner overfitting \cite{DBLP:conf/iclr/YinTZLF20} occurs when meta-learner overfit to insufficient training tasks, typically manifested by a meta-learner that can adapt quickly but still fails on the new task. Intuitively, learner overfitting is similar to overfitting in traditional learning, except that the smallest training unit is changed from samples to tasks \cite{DBLP:journals/corr/abs-2003-00804}. Those two forms of overfitting greatly degrade the generalization and robustness of meta-learning methods.


Data manipulation is a simple and efficient way to combat these overfitting issues ~\cite{DBLP:conf/icml/NiGSKG21,DBLP:conf/icml/YaoHZ0TZHL21}. There are two typical strategies: constructing mutually-exclusive tasks and conducting task-level augmentation. The previous one works on addressing memorization overfitting, where training tasks are independently assigned class labels to avoid the meta-learner handling tasks using rote memorization ~\cite{DBLP:conf/iclr/RaviL17,DBLP:conf/iclr/YinTZLF20}. The latter aims to alleviate learner overfitting by augmenting the training tasks. Typically, TaskAug ~\cite{DBLP:journals/corr/abs-2003-00804} rotates the image by a certain angle and treats it as a new class, thus increasing the diversity of the sampled tasks. To simultaneously alleviate these two forms of overfitting, the MetaMix ~\cite{DBLP:conf/icml/YaoHZ0TZHL21} linearly combines features and labels of samples to increase both the mutual-exclusivity and the diversity of training tasks. However, these regularization strategies based on data manipulation are manually designed for specific data or tasks, resulting in a lack of flexibility and generality in real-world applications:
\begin{itemize}
    \item Most strategies to increase task mutual-exclusivity are designed for classification tasks and are difficult to extend to other task scenarios \cite{DBLP:conf/iclr/YinTZLF20,DBLP:conf/cvpr/JamalQ19,DBLP:conf/icml/YaoHZ0TZHL21}.
    \item Our experiments in Sec. \ref{sec:exp_analysis} show that mutual-exclusivity is effective but short-lived, which means that simply increasing task mutual-exclusivity is not sufficient to combat memorization overfitting.
    \item Unless carefully designed based on the data, task augmentation is not always effective or even detrimental \cite{DBLP:conf/aistats/LiDB20,DBLP:conf/nips/YangZ020}. For example, Meta-MaxUp \cite{DBLP:conf/icml/NiGSKG21} systematically explores the effectiveness of various data-based augmentation strategies in meta-learning and reveals that \emph{shot augmentation} instead reduces the accuracy of the few-shot classification task. 
\end{itemize}

\IEEEpubidadjcol


Recent researches have focused on more flexible regularization strategies to address data-dependent limitations ~\cite{DBLP:conf/iclr/YinTZLF20}. For example, to combat memorization overfitting in regression tasks, MetaAug ~\cite{DBLP:conf/nips/RajendranIJ20} extends the mutually-exclusive property to the continuous sample space by perturbing labels with uniform noise. DropGrad ~\cite{DBLP:conf/accv/TsengCTLL020} provides a gradient regularization strategy where meta-gradients\footnote{For brevity, meta-learner related concepts are denoted by the prefix `meta-', e.g., `meta-gradient' for the gradient of meta-learner.} are randomly dropped to alleviate learner overfitting. Despite demonstrating promising results, it remains challenging to address both types of overfitting in a data-agnostic manner.

In this work, we improve generalization in meta-learning following the gradient regularization perspective and propose a data-independent \textbf{M}eta-\textbf{G}radient \textbf{Aug}mentation (\textbf{MGAug}). Considering that the meta-gradient required to update the meta-learner depends on the inner-loop fine-tuning of base learners, the key idea is to first solve the memorization overfitting issue in the inner loop by breaking the rote memorization state, and then yield  diversity gradients as the meta-gradient augmentation to alleviate learner overfitting in the outer loop. Specifically, breaking rote memories is achieved by pruning the base learner before each inner loop. To this end, we explore three different levels of pruning strategies, named \emph{random width pruning (WP)}, \emph{random parameter pruning (PP)}, and \emph{catfish pruning (CP)}\footnote{Our \textit{catfish pruning} is named after the ``catfish effect'', which forces sardines to reactivate by putting in catfish to avoid suffocation during transport. Here, sardines are base learner parameters that are forced to fine-tune on tasks to avoid rote states by \textit{catfish pruning}.}, respectively. The first two are based on random strategies and inspired by typical GradAug \cite{DBLP:conf/nips/YangZ020} and Dropout \cite{DBLP:journals/jmlr/SrivastavaHKSS14}, respectively, while CP is a newly proposed unstructured pruning strategy that measures a Meta-Memorization Carrying Amount (MMCA) score for each parameter and prunes those with high scores to achieve the most efficient memorization breaking. 

Once the rote memorization is removed, the pruned sub-network has to re-fine-tune the remaining parameters to handle new tasks, thus alleviating memorization overfitting. With different pruning rates, sub-networks produce gradients containing diverse task information as high-quality meta-gradient augmentation to ultimately reduce learner overfitting. The proposed MGAug is theoretically guaranteed by a PAC-Bayes-based generalization bound. In addition, we implemented a lightweight version (MGAug-MaxUp) inspired by the MaxUp strategy \cite{DBLP:journals/corr/abs-2002-09024}, as well as explored the plug-and-play property of MGAug as the trade-off between performance gains and computational resources. Extensive experimental results show that both MGAug and MGAug-MaxUp improve the generalization of various meta-learning baselines. 

The main contributions are summarized as follows:



\begin{itemize}
    \item We propose a novel data-agnostic meta-regularization via meta-gradient augmentation (MGAug), which can alleviate both memorization and learner overfitting in the two-loop meta-learning framework.

    \item We explore three pruning strategies to break rote memorization, including two existing random prunings and a new \emph{catfish pruning} that measures a Meta-Memorization Carrying Amount (MMCA) score for each parameter.

    \item We deduce a PAC-Bayes-based generalization bound as the theoretical guarantee for MGAug. In addition, we implement a lightweight extended version called MGAug-MaxUp that trades off performance and overhead.

    \item Experiments in both mutually-exclusive and non-mutually-exclusive tasks demonstrate that MGAug can be plug-and-played into most meta-learning baselines and significantly improve their performance.
\end{itemize}

\section{Related Work}
\subsection{Regularization}
Regularization techniques prevent the model from overfitting the training data and can be roughly divided into \emph{data augmentation}, \emph{label regularization}, and \emph{internal changes}. Data augmentation ~\cite{DBLP:journals/corr/abs-2006-06049,DBLP:conf/cvpr/YooAS20} and label regularization ~\cite{DBLP:conf/ijcnlp/GaoWHYN20,DBLP:conf/aistats/LiDB20} modify the input and labels respectively using various transformations (\textit{e.g.,} flipping and noise addition) to increase sample diversity. In contrast, internal variation emphasizes the diversity of parameters or connections, mostly independent of data or labels. The well-known Dropout ~\cite{DBLP:journals/jmlr/SrivastavaHKSS14} and its variants randomly remove some neurons to force the learner to capture more features. Shake-Shake ~\cite{DBLP:journals/corr/Gastaldi17} and ShakeDrop ~\cite{DBLP:journals/access/YamadaIAK19} are designed for a specific residual structure, giving different weights to each residual branch. The network pruning adopted in our MGAug belongs to an internal variation that prunes parameters to break rote memorization state. Another related work is GradAug ~\cite{DBLP:conf/nips/YangZ020}, which augments gradients to enrich network attention and improve generalization in conventional learning. The main difference is that our meta-gradient augmentation is naturally derived from pruned sub-networks in the meta-framework. 

\subsection{Meta regularization}
Meta regularization is specially designed for meta-learning to solve learner and memorization overfitting. For learner overfitting, well-designed task augmentation ~\cite{DBLP:journals/corr/abs-2003-00804} remains an effective solution by increasing task diversity. Meta-MaxUp ~\cite{DBLP:conf/icml/NiGSKG21} splits the meta-framework and further explores various data augmentation combinations. In contrast, memorization overfitting occurs in the inner loop with only a few updates, invalidating most conventional regularization strategies ~\cite{DBLP:conf/icml/YaoHZ0TZHL21,DBLP:conf/iclr/LeeNYH20}. Although constructing mutually-exclusive tasks ~\cite{DBLP:conf/iclr/LeeNYH20} shows promise against memorization overfitting, the task-dependent property makes it difficult to extend either diversity or mutual-exclusivity to regression and reinforcement learning scenarios ~\cite{DBLP:conf/iclr/YinTZLF20}. MetaPruning ~\cite{DBLP:conf/eccv/TianLYL20} ignores inner loops and improves meta-generalization through data-agnostic network pruning. Instead, we argue that redundant memories in inner loops are the key cause of memorization overfitting ~\cite{DBLP:journals/corr/abs-1907-07287,DBLP:journals/corr/abs-2109-14595}. With the same intuition, MR-MAML ~\cite{DBLP:conf/iclr/YinTZLF20} and TAML ~\cite{DBLP:conf/cvpr/JamalQ19} develop explicit meta-regularization terms to constrain the parameter scale and the base learner behavior, respectively. Unlike them, we directly break the rote memorization fetter via proposed \textit{catfish pruning} and alleviate learner overfitting using derived augmented meta-gradients, which can also be considered an enhanced DropGrad ~\cite{DBLP:conf/accv/TsengCTLL020}.

\section{Meta Learning}
Meta-learning is generally trained and tested on several tasks (here, we omit validation for brevity). To avoid confusion, we use the terms ``support set'' and ``query set'' to refer to training and test samples in a single task, leaving ``training set'' and ``testing set'' to the meta-learner ~\cite{DBLP:journals/corr/abs-2009-02653}. Given a set of training tasks $\{\mathcal{T}_t=(D^{s}_{t}, D^{q}_{t})\}$ sampled from the task distribution $p(\mathcal{T})$, where $D^{s}_{t} = (x^{s}_{t}, y^{s}_{t})$ is the support set containing support samples $x^{s}_{t}$ and corresponding labels $y^{s}_{t}$, and $D^{q}_{t} = (x^{q}_{t}, y^{q}_{t})$ is the query set containing query samples $x^{q}_{t}$ and labels $y^{q}_{t}$. The goal of meta-learning is to produce a base learner that can quickly handle new tasks $\mathcal{T}_{new}=(D^{s}_{new}, D^{q}_{new})$, i.e., fine-tune on the support data $(x^{s}_{new}, y^{s}_{new})$ and then accurately predict $y^{q}_{new}$ for $x^{q}_{new}$. Considering the difficulty of constructing a large number of routine tasks, meta-learning algorithms are usually validated on few-shot tasks. When applied to classification scenarios, this is commonly described as an $N$-way $K$-shot task, indicating $K$ samples in the support set, with class labels $y^s$, $y^q \in \{ 1, \dots, N \}$. In this way, each task consists of $KN$ support samples, and $K$ is usually small (typically $1$ or $5$).

\begin{figure*}[tp]
  \centering
  \includegraphics[width=0.7\linewidth]{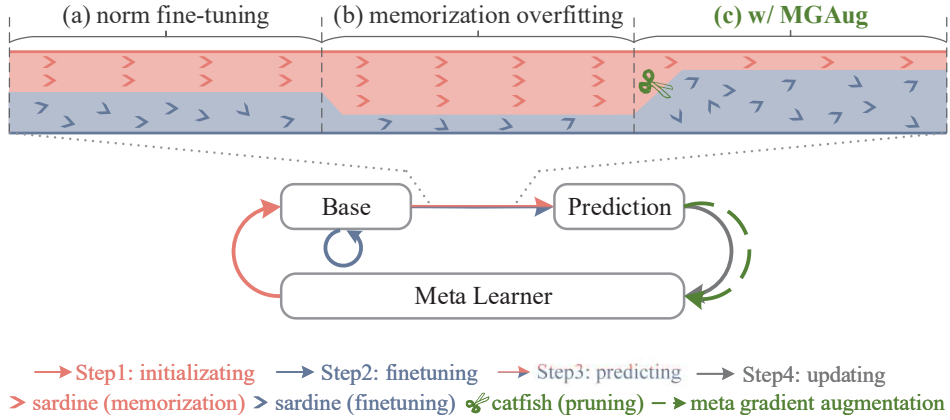}
  \caption{There are four steps in the meta-learner update, including initializing, fine-tuning, predicting, and updating. We zoom in on the inner-loop learning process and present three cases: (a) shows normal base learning, i.e., query samples are jointly predicted by initialization (meta memorization) and fine-tuning. (b) represents memorization overfitting, where query prediction mainly relies on rote memorization. (c) illustrates that the proposed \textit{catfish pruning} breaks meta memorization, forcing the base network to re-predict by fine-tuning and further deriving meta-gradient augmentation.}\label{fig:catfish}
\end{figure*}

\subsection{Two-loop meta-learning framework} Most meta-learning methods can be summarized into a two-loop framework ~\cite{DBLP:conf/nips/GoldblumFG20,DBLP:conf/icml/GoldblumRFNCG20}. The outer loop first involves sampling a batch of tasks $\{\mathcal{T}_t\}^T_{t=1} \sim p(\mathcal{T})$, and then updating meta-parameters based on \emph{feedback} derived from the inner loop over these tasks. In each inner loop, with the given meta-parameter $\omega$ and fine-tuning policy $\mathcal{F}$, the base parameters are updated from $\theta(\omega)$ to $\theta_{t}(\omega)$ on support samples $D^s_t$ for the $t$-th task:
\begin{equation}
\theta_t(\omega) = \mathcal{F}(\theta(\omega), D^s_t).
\label{eq:1_frame_inner}
\end{equation}

Afterwards, the \emph{feedback} is usually defined as the loss on query sample $D^q_t$ derived from the fine-tuned $\theta_t(\omega)$. Let $\mathcal{L}_{outer}$ denote the loss function in the outer loop, and meta-parameters are finally optimized by minimizing empirical risk, i.e., $\mathop{\arg\min}_{\omega} \sum_{t} \mathcal{L}_{outer}(\theta_t(\omega), D^q_t)$. Without loss of generality, using the gradient descent, meta-update in the $o$-th outer-loop can be further formalized as
\begin{equation}
\omega^{o} = \omega^{o-1} - \frac{\beta}{T} \sum_{t=1}^T \nabla_{\omega} \mathcal{L}_{outer}\left(\theta_t(\omega^{o-1}), D^q_t \right),
\label{eq:2_frame_outer}
\end{equation}
where $\beta$ is the meta-learning rate, $t$ and $T$ are the index and number of tasks, respectively. Note that different meta-parameter deployments $\theta(\omega)$ and fine-tuning algorithms $\mathcal{F}$ motivate different branches of meta-learning methods. 

The well-known gradient-based meta-learning (GBML) ~\cite{DBLP:conf/icml/LeeC18} takes $\omega$ as the initialization of base parameters $\theta$ and fine-tunes them by gradient descent, such as MAML ~\cite{DBLP:conf/icml/FinnAL17}, Meta-SGD ~\cite{DBLP:journals/corr/LiZCL17} etc. Formally, let $\mathcal{L}_{inner}$ denotes the loss in the inner loop, then $\mathcal{F} \triangleq \mathop{\arg\min}_{\theta} \mathcal{L}_{inner}(\theta(\omega), D^s_t)$. Similarly, after the $o$-th outer loop, the initialization and the $i$-th update of the base learner are as follows:
\begin{equation}
\theta^{o, i}_t = \theta^{o, i-1}_t - \alpha \nabla_{\theta} \mathcal{L}_{inner}(\theta^{o, i-1}_t, D^s_t), \dots, \theta^{o, 0}_t = \omega^o, 
\label{eq:3_frame_inner_gbml}
\end{equation}
where $\alpha$ is the base learning rate. Clearly, GBML expects to meta-learn good initialization to adapt quickly to unseen tasks. An alternative metric-based meta-learning (MBML) ~\cite{DBLP:conf/nips/SnellSZ17} meta-learns feature extractors and freezes them in inner loops, i.e., $\theta_{t}(\omega)=\theta(\omega)=\omega$. The query prediction is determined by the similarity between the support feature and the query feature, where the similarity is calculated by a non-parametric metric such as Euclidean distance or cosine distance. This paper focuses on these two meta-learning branches, but our MGAug can also be used for other branches and methods ~\cite{DBLP:conf/nips/GoldblumFG20} derived from this two-loop framework, such as R$2$-D$2$ ~\cite{DBLP:conf/iclr/BertinettoHTV19} and MetaOptNet ~\cite{DBLP:conf/cvpr/LeeMRS19}. 

\subsection{Two types of meta-overfitting} As mentioned, meta-overfitting consists of memorization and learner overfitting, where memorization overfitting is specific to the two-loop framework ~\cite{DBLP:conf/nips/RajendranIJ20,DBLP:conf/iclr/YinTZLF20}. We explain it by decomposing the meta-update into four steps in Fig. \ref{fig:catfish}: Step $1$, initialize base parameters $\theta(\omega)$ based on $\omega$; Step $2$, fine-tune $\theta(\omega)$ to $\theta_t(\omega)$ on the support set of the $t$-th task; Step $3$, predict the query sample based on $\theta_t(\omega)$ and calculate loss values; Step $4$, update $\omega$ once based on the average query error over a batch of tasks. Obviously, the query error is the key feedback for the update of meta-parameters and should be inferred jointly by meta-knowledge (i.e., $\omega$) and task-specific fine-tuning (see Fig. \ref{fig:catfish} (a)). Memorization overfitting occurs as $\omega$ is trained enough to directly memorize query predictions while ignoring fine-tuning (see Fig. \ref{fig:catfish} (b)), implying a degradation of rapid adaptability. Just as regular learners overfit training samples, another learner overfitting means that meta-learners may also overfit training tasks and fail to generalize to unseen novel tasks. Such learner overfitting can be naturally mitigated by task-level variants of regular regularizers ~\cite{DBLP:journals/corr/abs-2003-00804}.

\section{Meta-Gradient Augmentation}
This work aims to simultaneously mitigate these two types of meta-overfitting in a data-independent manner. The overall idea is to overcome memorization overfitting using network pruning and then alleviate learner overfitting with obtained augmented meta-gradients. Fig.\ref{fig:catfish} shows the illustration of our MGAug with the proposed \textit{catfish pruning}. In the inner loop, we first allow the base learner to fine-tune once normally and then compute the regular meta-gradient. To break the rote memorization state, \textit{catfish pruning} removes the parameters that carry the most meta-memorization to enforce base learner re-fine-tune to the support set (Fig. \ref{fig:catfish} (c)). Each pruning is like throwing a catfish into sardines (parameters of the base learner), resulting in different sub-networks and fine-tuning results. The higher the pruning rate, the more severe the memory breakage, and the more fine-tuning is required. After several independent pruning and fine-tuning stages, we obtain a set of augmented meta-gradients containing diversity task information, which are ultimately used to update the meta-learner as an augmentation of regular meta-gradients. Taking the GBML as an example, the rest of this section details the inner and outer loops using our MGAug.

\begin{figure*}[tp]
  \centering
  \includegraphics[width=0.7\linewidth]{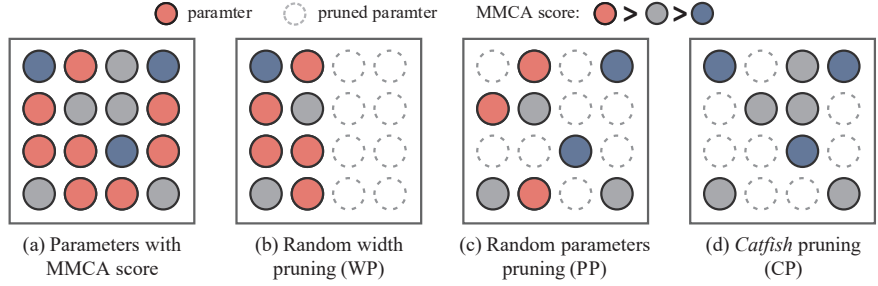}
  \caption{The comparison of different pruning strategies at 50\% pruning rate. (a) shows the unpruned network parameters, where the colors indicate the MMCA scores of the parameters. (b), (c), and (d) show the pruning results using WP, PP, and CP, respectively.}\label{fig:catfish}
\end{figure*}

\subsection{Inner-Loop with Network Pruning}
The inner-loop processes after the $o$-th outer loop, where the base parameters are initialized by the latest meta-learned parameters, i.e., $\theta(\omega) \triangleq \theta^{o, 0} = \omega^{o}$. Let $\mathcal{F}_\rho$ denote the pruning criterion. We can obtain the pruned sub-network parameters $\theta^{o, 0}_{\rho} = \mathcal{F}_\rho(\theta^{o, 0}, \rho)$ with a given pruning rate $\rho$, and rewrite the $i$-th update of the base learner on the $t$-th task in \eqref{eq:3_frame_inner_gbml} as
\begin{equation}
\theta^{o, i}_{\rho, t} = \theta^{o, i-1}_{\rho, t} -  \alpha \nabla \mathcal{L}_{inner}(\theta^{o,  i-1}_{\rho, t}, D^s_t).
\label{eq:4_catfish_inner}
\end{equation}
where $\alpha$ is the learning rate in the inner loop. For the pruning criterion $\mathcal{F}_\rho$, we explored three specific strategies whose intensity of memory breaking gradually increased at the same pruning rate. For brevity, we omit the superscripts of the inner and outer loops below.

\subsubsection{\textit{Random width pruning (WP)}}
WP is a structural pruning strategy that prunes the neurons in each layer of the network to meet the given pruning rate. Without loss of generality, we use the $l$-$th$ convolutional layer parameter $\theta_{(l)} \in \mathcal{R}^{d_{in} \times d_{out} \times k \times k}$ for illustration, where $k$ represents the convolution kernel size and $d_{in}$ and $d_{out}$ represent the number of input and output channels, respectively. For example, $d_{in}$ is the channel number of input images in the first layer, and $d_{out}$ is the number of corresponding convolution kernels. The number of trainable parameters in this layer is $n_{(l)} = d_{out} \times k \times k$. With the pruning rate $\rho \in [0,1]$, the parameter of the corresponding layer in sub-networks is $\theta_{\rho, (l)}$, where $\left| \theta_{\rho, (l)} \right| = n_{(l)}(1-\rho)$. Inspired by \cite{DBLP:conf/nips/YangZ020}, we sampled the first $(1-\rho) \times 100\%$ from the entire model as the sub-network, that is, $\theta_{\rho, (l)} \in \mathcal{R}^{(1-\rho) d_{in} \times (1-\rho) d_{out} \times k \times k}$. 


\subsubsection{\textit{Random parameter pruning (PP)}}
In contrast, PP is an unstructured pruning, where each parameter may be removed individually. Specifically, we introduce an indication mask $m \in \mathcal{R}^n$ consistent with the shape of base parameters $\theta$, where $n$ is the number of parameters and the value of $m$ is randomly selected from the set $\{0, 1\}$. A position with a value of $0$ in $m$ indicates that the corresponding parameter is pruned, otherwise it is retained. Afterwards, the pruned parameters can be expressed as $\theta_{\rho} = m \odot \theta$, where $\left| m \right| \leq n(1-\rho)$ and $\odot$ is the Hadamard product. Compared to WP, PP achieves internal changes at the parameter level, enabling more flexible memorization breaking.



\subsubsection{\textit{Catfish pruning (CP)}}
We further propose a task-oriented pruning criterion that enables stronger memorization breaking based on the current task and parameter state. Like PP, CP also needs an indicator mask $m$. The difference is that the mask value in CP reflects the amount of memories contained in each parameter rather than being randomly generated. To this end, we define the meta-memorization carrying amount and design a memorization-breaking pruning criterion.

\begin{definition}
\emph{\textbf{(Meta-Memorization Carrying Amount).} Let $\theta \in \mathbb{R}^n$ denote the base learner parameter and $e_{(j)}$ be the indicator vector for the $j$-th parameter $\theta_{(j)}$, whose value is zero everywhere except that index $j$ is one. Keeping everything else constant, we measure the query loss difference in the $t$-th task before and after pruning parameter $\theta_{(j)}$ to get the following Meta-Memorization Carrying Amount (\textbf{MMCA)}}:
\begin{equation}
 \begin{aligned}
    \text{MMCA}_{t, (j)} &\triangleq \Delta \mathcal{L}_{(j)}(\theta; D^q_t)\\
    &= \mathcal{L}\left(\mathbf{1} \odot \theta; D^q_t\right) - \mathcal{L}\left(\left(\mathbf{1}-e_{(j)}\right) \odot \theta; D^q_t\right),
 \end{aligned}
\label{eq:5_catfish_MMCA}
\end{equation} 
\emph{where $\mathbf{1}$ is the vector of dimension $m$ and $\odot$ denotes the Hadamard product.}
\end{definition}

MMCA essentially measures the sensitivity of parameter $\theta_{(j)}$ in solving task $t$. It is reasonable to represent the amount of memorization carried out here since the base parameters in each epoch are initialized by the meta parameters derived from the previous epoch. However, computing MMCA directly for each discrete parameter is prohibitively expensive as it requires $n + 1$ forward passes ($n$ is the number of parameters). Therefore, by relaxing the binary constraint on the indicator variable, we obtain an approximation of MMCA.

\begin{proposition} \emph{For any task  $\mathcal{T}_t=(D^{s}_{t}, D^{q}_{t})$ , the change in the loss function on $D^{q}_{t}$ before and after removing the $j$-th parameter $\theta_{(j)}$ can be approximated by}

\begin{equation}
\Delta \mathcal{L}_{(j)}(\theta; D^q_t) \approx \frac{\partial \mathcal{L}(\theta, D^q_t) }{\partial \theta_{(j)}} \times \theta_{(j)}.
\label{eq:6_catfish_MMCA_approx}
\end{equation}
\end{proposition}

We defer the proof to Appendix\ref{appendix1}. Similar approximation strategies ~\cite{DBLP:conf/icml/KohL17,DBLP:conf/iclr/LeeAT19} have also been used to measure the impact of a data point or connection on the loss. The key difference is that we leverage the fact that the query loss depends on meta-knowledge. Based on this MMCA score, we further compute the value of the binary mask $m$ by the designed memorization-breaking pruning criterion. Similar to PP, the parameters $\theta_{(l)}$ in the $l$-th layer are then pruned by $\theta_{\rho, (l)} = m_{(l)} \odot \theta_{(l)}$, where $|| m_{(l)} ||_0 / n_{(l)} \leq \rho$, where $n_{(l)} = |\theta_{(l)}|$ is the number of parameters in the $l$-th layer. 

\begin{definition}
\emph{\textbf{(Memorization-breaking pruning criterion).} Given a MMCA score mask, the parameters corresponding to the high score positions are removed to break the memorization state as much as possible, i.e., }

\begin{equation}
\begin{footnotesize}
m_{t, (j)} =
\begin{cases}
\ 0 & \text{if} \  |\text{MMCA}_{t, (j,l)} | \text{ is in the top-}\rho \text{\% largest value}\\
\ 1 & \text{otherwise}
\end{cases},
\end{footnotesize}
\label{eq:7_catfish_Cbreak}
\end{equation} \emph{where $(j,l)$ refers to the $j$-th parameter in the $l$-th layer. }
\end{definition}

Such we obtain a layer-level binary mask and can further compute the sub-network predictions via forwarding propagation. During backward propagation, we apply the same mask to the gradient so that the pruned parameters are no longer updated while the others are trained normally.

\begin{figure}[tp]
    \centering
    \includegraphics[width=0.75\linewidth]{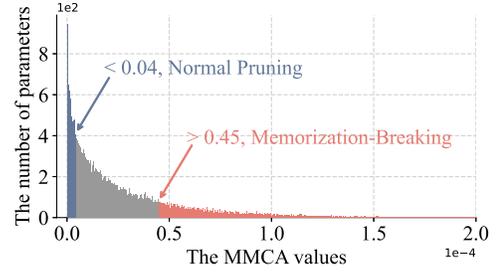}
    \caption{MMCA distribution of the last layer in Conv-4.}
    \label{fig:MMCA}
\end{figure}

\begin{remark}
    \emph{Our catfish pruning breaks rote memorization to combat memorization overfitting and generates augmented meta-gradients containing diversity task information, with two differences from connection-sensitivity-based network pruning methods ~\cite{DBLP:conf/icml/KohL17,DBLP:conf/nips/TanakaKYG20,DBLP:conf/iclr/FrankleD0C21}. One is that \textit{catfish pruning} prunes in the inner loop, so the criterion is based on the query set rather than the regular training samples ~\cite{DBLP:conf/iclr/WangZG20}. Another essential difference is that normal pruning usually removes insensitive parameters for accuracy preservation, which is even the exact opposite of our memorization-breaking criterion ~\cite{DBLP:conf/iclr/FrankleC19,wang2022snip,DBLP:conf/cvpr/ChenFC0ZCW21}. We highlight this difference in Fig. \ref{fig:MMCA} by visualizing the MMCA distribution of the last layer parameters in the Conv-$4$ backbone with $20\%$ pruning rate. }
\end{remark}

\subsection{Outer-loop with Augmented Meta-Gradients}
We now obtain two group gradients with respect to the meta-parameters $w^o$: one is the (normal) meta-gradients obtained by backward-propagation on the full network, and the other is derived from several sub-networks. The former retains full meta-knowledge to speed the training process and avoid underfitting at the early learning stage. The latter is the meta-gradient augmentation resulting from structural perturbations. The meta-parameters are finally updated by accumulating these two-group gradients and formalized as
\begin{equation}
    \begin{aligned}
        \omega^{o+1} = \omega^o - \frac{\beta}{T} \sum_{t=1}^T &\Big(\nabla_{\omega} \mathcal{L}_{outer}(\theta_t(\omega), D^q_t) \\ 
        &+ \sum_{u=1}^U \nabla_{\omega}\mathcal{L}_{outer}(\theta_{u, t}(\omega), D^{q}_{t}) \Big),
        \label{eq:8_catfish_outer}  
 	\end{aligned}
\end{equation}
where $U$ is a hyper-parameter representing the number of sub-networks in each task. Since the meta-learner is trained with shared initialization parameters, it naturally shares diversity attention across all sub-networks, which is the key to combating learner overfitting. The entire procedure for the two-loop framework with MGAug is provided in Algorithm \ref{alg:framework_our}.

\begin{remark}
    \emph{MGAug is designed for the two-loop meta-framework and is quite different from the Dropout-style approaches \cite{DBLP:journals/jmlr/SrivastavaHKSS14,DBLP:conf/accv/TsengCTLL020}. The former prunes base parameters before each inner loop in order to break the rote memorization state, while the latter directly prunes meta-parameters in the outer loop, essentially to alleviate learner overfitting.}
\end{remark}

\begin{remark}
    \emph{The augmented meta-gradients derived from CP in MGAug are based on the task response rather than directly changing training tasks, which is also essentially different from the stochastic gradient noise strategy ~\cite{DBLP:conf/accv/TsengCTLL020}. The advantages of similar self-guided augmentation have been validated in traditional learning ~\cite{DBLP:conf/nips/YangZ020}.}
\end{remark}

\begin{algorithm}[tb]
\caption{Two-loop framework with MGAug.}
\label{alg:framework_our}
\small
\textbf{Require}: Meta parameters $\omega$, base parameters $\theta$, fine-tuning algorithm $\mathcal{F}$, meta learning rate $\beta$, task distribution $p(\mathcal{T})$, pruning method $\mathcal{F}_\rho$, pruning rate range $[\rho_{min}, \rho_{max})$, and the number of sub-networks $U$.
\begin{algorithmic}[1] 
\STATE Initialize $\omega$, $\theta(\omega)$;
\WHILE{not done} 
\STATE Sample batch of tasks $\{{\mathcal{T}_t = (D^{s}_{t}, D^{q}_{t})\}^{T}_{t=1}} \sim p(\mathcal{T})$.
\FOR {$t = 1, \dots, T$}
\STATE Fine-tune base parameters on $\mathcal{T}_t$, obtain new base parameters $\theta_t(\omega) = \mathcal{F}(\theta(\omega), D^{s}_{t})$.
\STATE Compute full meta-gradients $g_t^{ori} = \nabla_{\omega}\mathcal{L}(\theta_t(\omega), D^{q}_{t})$
\FOR { $u = 1, \dots, U$} 
\STATE Randomly set pruning rate $\rho_u \in [\rho_{min}, \rho_{max})$
\STATE Pruning to obtain base sub-parameters $\theta_u(\omega)$ = $\mathcal{F}_\rho(\theta(\omega),\rho_u, g_t)$
\STATE Fine-tune sub-parameters on $\mathcal{T}_t$, obtain new sub-parameters $\theta_{u, t}(\omega) = \mathcal{F}(\theta_u(\omega),D^{s}_{t})$.
\ENDFOR
\IF {MGAug-MaxUp}
\STATE Compute $g^{aug}_t = \nabla_{\omega} \max_{u \in U} \mathcal{L}(\theta_{u, t}(\omega), D^{q}_{t})$
\STATE Compute $g_t = \max(g^{aug}_t, g^{ori}_t)$

\ELSE
\STATE Compute $g^{aug}_t = \sum_u \nabla_{\omega}\mathcal{L}(\theta_{u, t}(\omega), D^{q}_{t})$
\STATE Compute $g_t = g^{aug}_t + g^{ori}_t$
\ENDIF
\ENDFOR
\STATE Update meta parameters: $\omega \gets \omega - \frac{\beta}{T}\sum_t g_t$
\ENDWHILE
\end{algorithmic}
\end{algorithm}

\subsection{MGAug-MaxUp: A Lightweight Version}
As an alternative, we further extended a lightweight version, noted as MGAug-MaxUp. Inspired by MaxUp \cite{DBLP:journals/corr/abs-2002-09024}, MGAug-MaxUp only updates meta-learner by back-propagating in the network with the largest query loss instead of all networks. In this way, the update of the outer loop becomes:

\begin{equation}
    \begin{aligned}
        \omega^{o+1} = \omega^o - \frac{\beta}{T} \sum_{t=1}^T \Big(& \nabla_{\omega} \max \big(\mathcal{L}_{outer}(\theta_t(\omega), D^q_t), \\
        & \{\mathcal{L}_{outer}(\theta_{u, t}(\omega), D^{q}_{t})\}_{u=1}^{U} \big) \Big),
    \label{eq:9_maxcatfish}
    \end{aligned}
\end{equation}
where $\max(\cdot)$ is the maximum function, which itself does not back-propagate during training but is simply the gradient of the worst copy. As a tradeoff between performance gains and computational costs, MGAug-MaxUp can be easily deployed in resource-limited scenarios.


\subsection{A PAC-Bayes Generalization Bound}
We provide a PAC-Bayes-based generalization bound \cite{DBLP:journals/corr/McAllester13} for the two-loop meta-learning framework with inner-loop pruning, which can theoretically guarantee the performance of our MGAug. We simplify the analysis by pruning a sub-network for each task (i.e., $U=1$). Following the meta-learning PAC-Bayes framework~\cite{DBLP:conf/icml/AmitM18}, let $\mathcal{P}$ and $\mathcal{Q}$ be the hyper-prior and hyper-posterior of the meta-learner, and assume that loss function is bounded to the interval $[0,1]$. For a given pruning rate $\rho\in[0,1]$ and initial parameters $\Theta_i$ of the base learner on task $i$, we take the pruned parameters $Q_{\rho,\Theta_i}$ as posterior distribution and the corresponding $Q_{\rho,0} \sim \mathcal{Q}$ as the prior distribution. 

\begin{theorem} \textbf{\emph{(Meta-learning PAC-Bayes bound with inner-loop pruning).}} \label{theorem1} Let $er(\mathcal{Q})$ and $\hat{er}(Q_{\rho,\Theta}, \mathcal{T})$ be the expected and empirical errors in meta-learning, and let $m_i$ denote the number of samples in the $i$-$th$ task. Then for any $\delta \in (0,1]$ the following inequality holds uniformly for all hyper-posterior distributions $\mathcal{Q}$ with probability at least $1-\delta$,

\begin{footnotesize}
\begin{equation*}
\begin{aligned}
er(\mathcal{Q}) \leq \frac{1}{T}\sum_{i=1}^{T} \underset{Q_{\rho, 0} \sim \mathcal{Q}}{\mathrm{E}} \hat{er}_i(Q_{\rho,\Theta_i}, \mathcal{T}_i) + \sqrt{\frac{D(\mathcal{Q}\Vert \mathcal{P}) + \log \frac{2T}{\delta}}{2(T-1)}}\\ 
+ \frac{1}{T}\sum_{i=1}^{T}\sqrt{\frac{D(\mathcal{Q}\Vert \mathcal{P}) + \log \frac{2Tm_i}{\delta} + \frac{1-\rho}{2}\Vert \Theta_i \Vert^2}{2(m_i-1)}}. 
\label{eq:9_bound}
\end{aligned}
\end{equation*}
\end{footnotesize}
\end{theorem}
The expected error is bounded by the empirical multi-task error plus two complexity terms. The first is the average of task-complexity terms for observed tasks, and the second is the environment-complexity term \cite{DBLP:conf/icml/AmitM18}. MGAug prunes parameters in the inner loop, reducing the complexity cost of base learners by a factor of $1-\rho$ and further reducing the task-complexity terms. On the other hand, complexity terms are independent on the pruning criterion. A good criterion can improve generalization by minimizing the increase in empirical error \cite{DBLP:journals/corr/abs-2205-07320}. Appendix\ref{appendix2} provides proof of Theorem \ref{theorem1}.

\section{Experiments}   \label{sec:Exp}

This section presents extensive experimental results of MGAug and its lightweight versions on multiple public benchmarks. The remainder of the experiments are organized as follows: Subsection \ref{sec:exp_setting} lists the basic experimental setups, including datasets, backbones, hyper-parameters, etc. Subsection \ref{sec:exp_sota} shows the generalization improvement brought by MGAug and the comparison with state-of-the-art methods over various meta-learning instances. Subsection \ref{sec:exp_analysis} explores the reasons why MGAug works well by analyzing its behaviors from different perspectives. Subsections \ref{sec:exp_hyperparams} and \ref{sec:exp_other} investigate the robustness of MGAug for different hyper-parameters and scenarios, respectively.

\begin{table*}[t]
    \centering
    \caption{Classification accuracy of different regularization methods for the ProtoNet baseline on CUB tasks. \label{tab:result_proto_cub}}
    \resizebox{1.\linewidth}{!}{
        \begin{tabular}{*{7}{l lll lll}}
            \toprule
            \multirow{2}*{} &  \multicolumn{3}{c}{5-way 1-shot} & \multicolumn{3}{c}{5-way 5-shot}  \\
            \cmidrule(lr){2-4}\cmidrule(lr){5-7}
            & Conv-$4$     & ResNet-$10$     & ResNet-$18$     & Conv-$4$     & ResNet-$10$     & ResNet-$18$\\
            \midrule
            ProtoNet                    & 45.26{\tiny$\pm$0.90}		    & 50.29{\tiny$\pm$0.89}         & 57.87{\tiny$\pm$1.03}           & 66.25{\tiny$\pm$0.71}		  & 71.41{\tiny$\pm$0.67}	 	  & 70.92{\tiny$\pm$0.66}            \\
            + Aug                       & 55.18{\tiny$\pm$0.97}		    & 71.61{\tiny$\pm$0.87}         & 74.25{\tiny$\pm$0.96}           & 75.93{\tiny$\pm$0.67}		  & 84.26{\tiny$\pm$0.53}	 	  & 85.87{\tiny$\pm$0.50}            \\
            \midrule
            + TaskAug \cite{DBLP:journals/corr/abs-2003-00804}  & 57.64{\tiny$\pm$0.97} (2.46$\uparrow$) & 73.44{\tiny$\pm$0.89} (1.83$\uparrow$) & 76.31{\tiny$\pm$0.95} (2.06$\uparrow$)   & 78.21{\tiny$\pm$0.65} (2.28$\uparrow$) & 85.78{\tiny$\pm$0.50} (1.52$\uparrow$) & 87.63{\tiny$\pm$0.46} (1.76$\uparrow$)  \\
            + Meta-MaxUp \cite{DBLP:conf/icml/NiGSKG21}         & 59.79{\tiny$\pm$0.90} (4.61$\uparrow$) & 75.20{\tiny$\pm$0.85} (3.59$\uparrow$) & 77.03{\tiny$\pm$0.92} (2.78$\uparrow$)   & 78.86{\tiny$\pm$0.57} (2.93$\uparrow$) & 86.02{\tiny$\pm$0.42} (1.76$\uparrow$) & 87.92{\tiny$\pm$0.44} (2.05$\uparrow$)  \\
            \midrule
            + \textbf{MGAug-WP}         & \underline{62.35{\tiny$\pm$0.97}} (7.17$\uparrow$) & 76.30{\tiny$\pm$0.88} (4.69$\uparrow$) & 77.74{\tiny$\pm$0.88} (3.49$\uparrow$)   & 79.57{\tiny$\pm$0.62} (3.64$\uparrow$) & 85.19{\tiny$\pm$0.49} (0.93$\uparrow$) & 87.74{\tiny$\pm$0.46} (1.87$\uparrow$)  \\
            + \textbf{MGAug-PP}         & 61.28{\tiny$\pm$0.98} (6.10$\uparrow$) & 76.01{\tiny$\pm$0.88} (4.40$\uparrow$) & \underline{78.42{\tiny$\pm$0.93}} (4.17$\uparrow$)   & 79.06{\tiny$\pm$0.61} (3.13$\uparrow$) & 85.74{\tiny$\pm$0.49} (1.48$\uparrow$) & \underline{88.72{\tiny$\pm$0.46}} (2.85$\uparrow$)  \\
            + \textbf{MGAug-CP}         & \textbf{63.00{\tiny$\pm$0.95} (7.82$\uparrow$)} & \textbf{78.57{\tiny$\pm$0.92}} (6.96$\uparrow$) & 77.22{\tiny$\pm$0.95} (2.97$\uparrow$)   & \textbf{80.33{\tiny$\pm$0.62}} (4.40$\uparrow$) & \underline{87.09{\tiny$\pm$0.46}} (2.83$\uparrow$) & \textbf{89.68{\tiny$\pm$0.45}} (3.81$\uparrow$)  \\
            \midrule
            + \textbf{MGAug-MaxUp-WP}   & 60.53{\tiny$\pm$0.97} (5.35$\uparrow$) & 73.92{\tiny$\pm$0.85} (2.31$\uparrow$) & \textbf{78.55{\tiny$\pm$0.94}} (4.30$\uparrow$)   & 79.51{\tiny$\pm$0.63} (3.58$\uparrow$) & 85.39{\tiny$\pm$0.52} (1.13$\uparrow$) & 88.30{\tiny$\pm$0.46} (2.43$\uparrow$)  \\
            + \textbf{MGAug-MaxUp-PP}   & 60.61{\tiny$\pm$0.98} (5.43$\uparrow$) & 74.19{\tiny$\pm$0.92} (2.58$\uparrow$) & 77.77{\tiny$\pm$0.96} (3.52$\uparrow$)   & 79.81{\tiny$\pm$0.62} (3.88$\uparrow$) & 86.27{\tiny$\pm$0.48} (2.01$\uparrow$) & 88.59{\tiny$\pm$0.43} (2.72$\uparrow$)  \\
            + \textbf{MGAug-MaxUp-CP}   & 61.18{\tiny$\pm$0.97} (6.00$\uparrow$) & \underline{76.77{\tiny$\pm$0.87}} (5.16$\uparrow$) & 75.32{\tiny$\pm$0.92} (1.07$\uparrow$)   & \underline{79.83{\tiny$\pm$0.60}} (3.90$\uparrow$) & \textbf{87.66{\tiny$\pm$0.48}} (3.40$\uparrow$) & 88.66{\tiny$\pm$0.43} (2.79$\uparrow$)  \\
            \bottomrule
        \end{tabular}
    }
\end{table*}


\begin{table*}[h]
    \centering
    \caption{Classification accuracy of different regularization methods for the ProtoNet baseline on \emph{mini}-ImageNet tasks. \label{tab:result_proto_mini}}
    \begin{tabular}{*{7}{l lll lll}}
        \toprule
        \multirow{2}*{} &  \multicolumn{3}{c}{5-way 1-shot} & \multicolumn{3}{c}{5-way 5-shot}  \\
        \cmidrule(lr){2-4}\cmidrule(lr){5-7}
        & Conv-$4$     & ResNet-$10$     & ResNet-$18$     & Conv-$4$     & ResNet-$10$     & ResNet-$18$\\
        \midrule
        ProtoNet                    & 31.37{\tiny$\pm$0.62}		 & 43.54{\tiny$\pm$0.80}         & 45.69{\tiny$\pm$0.82}           & 65.10{\tiny$\pm$0.72}		 & 63.29{\tiny$\pm$0.67}	 	   & 61.58{\tiny$\pm$0.68}  \\
        + Aug                       & 44.79{\tiny$\pm$0.82}		 & 51.65{\tiny$\pm$0.83}         & 57.06{\tiny$\pm$0.91}           & 65.98{\tiny$\pm$0.72}		 & 74.02{\tiny$\pm$0.65}	 	   & 73.54{\tiny$\pm$0.66}  \\
        \midrule
        + TaskAug \cite{DBLP:journals/corr/abs-2003-00804}  & 42.55{\tiny$\pm$0.78} (2.24$\downarrow$) & 56.33{\tiny$\pm$0.89} (4.68$\uparrow$) & 59.27{\tiny$\pm$0.92} (2.21$\uparrow$) & 63.97{\tiny$\pm$0.76} (2.01$\downarrow$) & 74.79{\tiny$\pm$0.65} (0.77$\uparrow$) & 74.02{\tiny$\pm$0.67} (0.48$\uparrow$)  \\
        + Meta-MaxUp \cite{DBLP:conf/icml/NiGSKG21}         & 45.47{\tiny$\pm$0.82} (0.68$\uparrow$) & 57.52{\tiny$\pm$0.82} (5.87$\uparrow$) & 60.20{\tiny$\pm$0.88} (3.14$\uparrow$)   & 66.22{\tiny$\pm$0.72} (0.24$\uparrow$) & 74.83{\tiny$\pm$0.60} (0.81$\uparrow$) & 74.83{\tiny$\pm$0.66} (1.29$\uparrow$)  \\
        \midrule
        + \textbf{MGAug-WP}         & 46.79{\tiny$\pm$0.84} (2.00$\uparrow$) & 57.87{\tiny$\pm$0.86} (6.22$\uparrow$) & \textbf{59.90{\tiny$\pm$0.93}} (2.84$\uparrow$)   & 67.31{\tiny$\pm$0.70} (1.33$\uparrow$) & 74.26{\tiny$\pm$0.64} (0.24$\uparrow$) & 74.67{\tiny$\pm$0.62} (1.13$\uparrow$)  \\
        + \textbf{MGAug-PP}         & 47.68{\tiny$\pm$0.83} (2.89$\uparrow$) & 55.26{\tiny$\pm$0.86} (3.61$\uparrow$) & 59.18{\tiny$\pm$0.92} (2.12$\uparrow$)   & 67.84{\tiny$\pm$0.71} (1.86$\uparrow$) & \underline{75.53{\tiny$\pm$0.64}} (1.51$\uparrow$) & 75.51{\tiny$\pm$0.63} (1.97$\uparrow$)  \\
        + \textbf{MGAug-CP}         & \textbf{48.77{\tiny$\pm$0.86}} (3.98$\uparrow$) & \textbf{58.32{\tiny$\pm$0.86}} (6.67$\uparrow$) & 58.90{\tiny$\pm$0.91} (1.84$\uparrow$)   & \underline{67.99{\tiny$\pm$0.73}} (2.01$\uparrow$) & \textbf{75.77{\tiny$\pm$0.63}} (1.75$\uparrow$) & \textbf{76.05{\tiny$\pm$0.63}} (2.51$\uparrow$) \\
        \midrule
        + \textbf{MGAug-MaxUp-WP}   & 45.60{\tiny$\pm$0.80} (0.81$\uparrow$) & 57.15{\tiny$\pm$0.85} (5.50$\uparrow$) & 58.82{\tiny$\pm$0.93} (1.76$\uparrow$)   & \textbf{68.39{\tiny$\pm$0.70}} (2.41$\uparrow$) & 75.01{\tiny$\pm$0.65} (0.99$\uparrow$) & 74.91{\tiny$\pm$0.67} (1.37$\uparrow$)\\
        + \textbf{MGAug-MaxUp-PP}   & 46.35{\tiny$\pm$0.81} (1.56$\uparrow$) & 56.88{\tiny$\pm$0.88} (5.23$\uparrow$) & \underline{59.36{\tiny$\pm$0.92}} (2.30$\uparrow$)   & 67.54{\tiny$\pm$0.71} (1.56$\uparrow$) & 74.74{\tiny$\pm$0.62} (0.72$\uparrow$) & 75.37{\tiny$\pm$0.63} (1.83$\uparrow$)  \\
        + \textbf{MGAug-MaxUp-CP}   & \underline{47.73{\tiny$\pm$0.83}} (2.94$\uparrow$) & \underline{58.10{\tiny$\pm$0.85}} (6.45$\uparrow$)  & 59.05{\tiny$\pm$0.92} (1.99$\uparrow$)   & 66.71{\tiny$\pm$0.74} (0.73$\uparrow$) & 74.54{\tiny$\pm$0.64} (0.52$\uparrow$) & \underline{75.72{\tiny$\pm$0.63}} (2.18$\uparrow$)  \\
        \bottomrule
    \end{tabular}
\end{table*}


\begin{figure*}[t]
  \centering
  \includegraphics[width=0.92\textwidth]{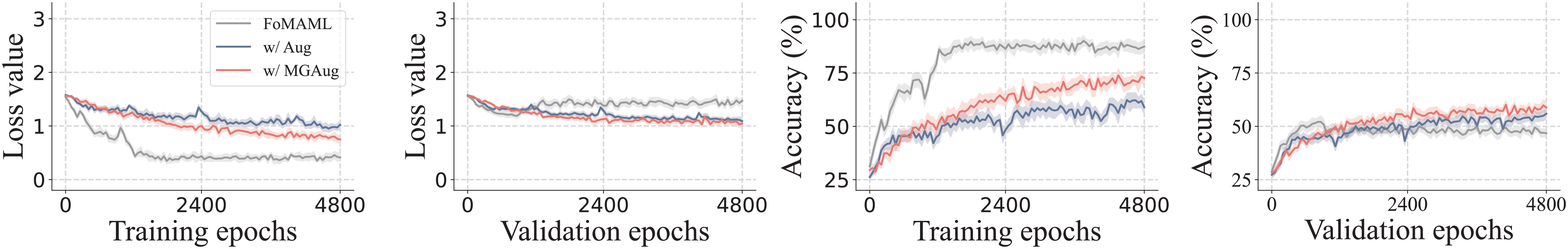}
  \includegraphics[width=0.92\textwidth]{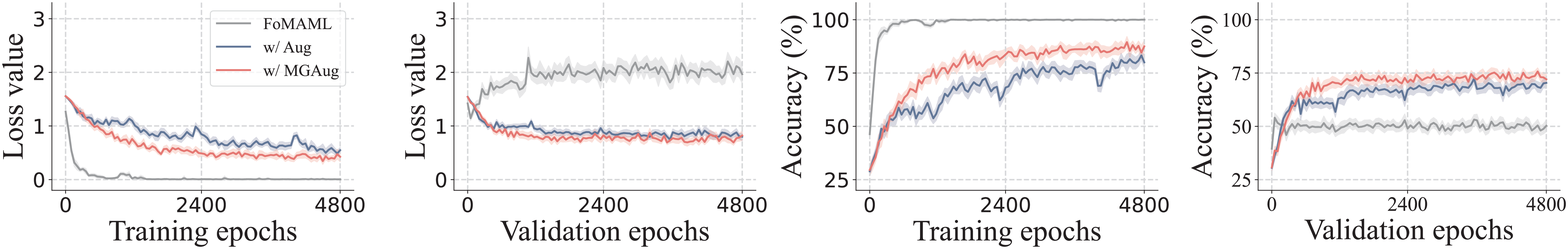}
  \caption{Loss and accuracy curves of FoMAML baseline, Aug, and MGAug using ResNet-$10$ backbone on \emph{mini}-Imagenet (top) and CUB (bottom). From left to right, the first two are the loss curves during training and validation, respectively, and the last two are the corresponding accuracy curves.}\label{fig:loss}
\end{figure*}

\begin{table*}[h]
    \centering
    \caption{Classification accuracy of different regularization methods for the FoMAML baseline on CUB tasks. \label{tab:result_fomaml_cub}}
    \begin{tabular}{*{7}{l lll lll}}
        \toprule
        \multirow{2}*{} &  \multicolumn{3}{c}{5-way 1-shot} & \multicolumn{3}{c}{5-way 5-shot}  \\
        \cmidrule(lr){2-4}\cmidrule(lr){5-7}
        & Conv-$4$             & ResNet-$10$         & ResNet-$18$            & Conv-$4$     & ResNet-$10$     & ResNet-$18$\\
        \midrule
        FoMAML                      & 53.21{\tiny$\pm$0.48}      & 54.69{\tiny$\pm$0.51}   & 55.60{\tiny$\pm$0.52}       & 69.08{\tiny$\pm$0.39}  & 64.49{\tiny$\pm$0.43}  & 65.84{\tiny$\pm$0.43}  \\
        + Aug                       & 53.98{\tiny$\pm$0.48}      & 68.14{\tiny$\pm$0.50}   & 70.67{\tiny$\pm$0.51}       & 73.66{\tiny$\pm$0.36}  & 78.17{\tiny$\pm$0.36}  & 82.01{\tiny$\pm$0.34}  \\
        \midrule
        + MR \cite{DBLP:conf/iclr/YinTZLF20}         & 56.11{\tiny$\pm$0.54} (2.13$\uparrow$) & 69.54{\tiny$\pm$0.52} (1.40$\uparrow$) & 70.69{\tiny$\pm$0.31} (0.02$\uparrow$) & 74.73{\tiny$\pm$0.45} (1.07$\uparrow$)  & 79.26{\tiny$\pm$0.40} (1.09$\uparrow$) & 82.35{\tiny$\pm$0.57} (0.34$\uparrow$) \\
        + TAML \cite{DBLP:conf/cvpr/JamalQ19}        & 55.64{\tiny$\pm$0.85} (1.66$\uparrow$) & 70.22{\tiny$\pm$0.81} (2.08$\uparrow$) & 71.42{\tiny$\pm$0.60} (0.75$\uparrow$) & 75.39{\tiny$\pm$0.74} (1.73$\uparrow$)  & 78.72{\tiny$\pm$0.69} (0.55$\uparrow$) & 83.49{\tiny$\pm$0.55} (1.48$\uparrow$) \\
        + GradDrop \cite{DBLP:conf/accv/TsengCTLL020}& 55.39{\tiny$\pm$0.51} (1.41$\uparrow$) & 69.03{\tiny$\pm$0.52} (0.89$\uparrow$) & 71.29{\tiny$\pm$0.73} (0.62$\uparrow$) & 74.88{\tiny$\pm$0.38} (1.22$\uparrow$)  & 80.59{\tiny$\pm$0.34} (2.42$\uparrow$) & 82.46{\tiny$\pm$0.36} (0.45$\uparrow$) \\
        + MetaMix \cite{DBLP:conf/icml/YaoHZ0TZHL21} & \underline{57.53{\tiny$\pm$0.53}} (3.55$\uparrow$) & 70.38{\tiny$\pm$0.44} (2.24$\uparrow$) & \underline{72.97{\tiny$\pm$0.68}} (2.30$\uparrow$) & 76.24{\tiny$\pm$0.40} (2.58$\uparrow$)  & 80.83{\tiny$\pm$0.35} (2.66$\uparrow$) & 83.80{\tiny$\pm$0.38} (1.79$\uparrow$) \\
        \midrule   
        + \textbf{MGAug-WP}         & 56.85{\tiny$\pm$0.50} (2.87$\uparrow$) & 70.73{\tiny$\pm$0.46} (2.59$\uparrow$) & 72.10{\tiny$\pm$0.51} (1.43$\uparrow$) & 75.89{\tiny$\pm$0.42} (2.23$\uparrow$) & \underline{81.53{\tiny$\pm$0.44}} (3.36$\uparrow$) & 83.26{\tiny$\pm$0.30} (1.25$\uparrow$) \\
        + \textbf{MGAug-PP}         & 55.35{\tiny$\pm$0.47} (1.37$\uparrow$) & 71.17{\tiny$\pm$0.47} (3.03$\uparrow$) & 72.25{\tiny$\pm$0.47} (1.58$\uparrow$) & 76.16{\tiny$\pm$0.36} (2.50$\uparrow$) & 81.47{\tiny$\pm$0.35} (3.30$\uparrow$) & 83.92{\tiny$\pm$0.30} (1.91$\uparrow$) \\
        + \textbf{MGAug-CP}         & \textbf{58.19{\tiny$\pm$0.49}} (4.21$\uparrow$) & \textbf{72.14{\tiny$\pm$0.51}} (4.00$\uparrow$) & \textbf{73.59{\tiny$\pm$0.49}} (2.92$\uparrow$) & 76.28{\tiny$\pm$0.35} (2.62$\uparrow$) & \textbf{81.97{\tiny$\pm$0.33}} (3.80$\uparrow$) & \textbf{84.23{\tiny$\pm$0.32}} (2.22$\uparrow$) \\
        \midrule
        + \textbf{MGAug-MaxUp-WP}   & 56.53{\tiny$\pm$0.53} (2.55$\uparrow$) & 70.69{\tiny$\pm$0.47} (2.55$\uparrow$) & 71.37{\tiny$\pm$0.42} (0.70$\uparrow$) & 76.05{\tiny$\pm$0.33} (2.39$\uparrow$) & 80.92{\tiny$\pm$0.41} (2.75$\uparrow$) & 82.63{\tiny$\pm$0.31} (0.62$\uparrow$) \\
        + \textbf{MGAug-MaxUp-PP}   & 56.88{\tiny$\pm$0.50} (2.90$\uparrow$) & 71.32{\tiny$\pm$0.48} (3.18$\uparrow$) & 72.64{\tiny$\pm$0.41} (1.97$\uparrow$) & \underline{76.42{\tiny$\pm$0.35}} (2.76$\uparrow$) & 81.40{\tiny$\pm$0.43} (3.23$\uparrow$) & 83.32{\tiny$\pm$0.33} (1.31$\uparrow$) \\
        + \textbf{MGAug-MaxUp-CP}   & 57.06{\tiny$\pm$0.50} (3.08$\uparrow$) & \underline{71.61{\tiny$\pm$0.51}} (3.47$\uparrow$) & 72.58{\tiny$\pm$0.41} (1.91$\uparrow$) & \textbf{77.03{\tiny$\pm$0.31}} (3.37$\uparrow$) & 81.44{\tiny$\pm$0.43} (3.27$\uparrow$) & \underline{84.01{\tiny$\pm$0.28}} (2.00$\uparrow$)  \\
        \bottomrule
    \end{tabular}
\end{table*}

\begin{table*}[h]
    \centering
    \caption{Classification accuracy of different regularization methods for the FoMAML baseline on \emph{mini}-ImageNet tasks. \label{tab:result_fomaml_mini}}
    \begin{tabular}{*{7}{l lll lll}}
        \toprule
        \multirow{2}*{} &  \multicolumn{3}{c}{5-way 1-shot} & \multicolumn{3}{c}{5-way 5-shot}  \\
        \cmidrule(lr){2-4}\cmidrule(lr){5-7}
        & Conv-$4$             & ResNet-$10$         & ResNet-$18$            & Conv-$4$     & ResNet-$10$     & ResNet-$18$\\
        \midrule
        FoMAML                      & 43.50{\tiny$\pm$0.39}  & 46.79{\tiny$\pm$0.46} & 42.87{\tiny$\pm$0.42}    & 59.28{\tiny$\pm$0.38} & 57.86{\tiny$\pm$0.39} & 62.82{\tiny$\pm$0.38} \\
        + Aug                       & 44.41{\tiny$\pm$0.40}  & 52.16{\tiny$\pm$0.47} & 51.83{\tiny$\pm$0.46}    & 60.82{\tiny$\pm$0.38} & 66.93{\tiny$\pm$0.38} & 66.24{\tiny$\pm$0.37} \\
        \midrule
        + MR \cite{DBLP:conf/iclr/YinTZLF20}         & 44.57{\tiny$\pm$0.41} (0.16$\uparrow$) & 52.65{\tiny$\pm$0.56} (0.49$\uparrow$)   & 52.47{\tiny$\pm$0.51} (0.64$\uparrow$) & 61.15{\tiny$\pm$0.38} (0.33$\uparrow$) & 68.20{\tiny$\pm$0.51} (1.27$\uparrow$) & 66.71{\tiny$\pm$0.38} (0.47$\uparrow$) \\
        + TAML \cite{DBLP:conf/cvpr/JamalQ19}        & 45.72{\tiny$\pm$0.30} (1.31$\uparrow$) & 51.78{\tiny$\pm$0.53} (0.38$\downarrow$) & 53.19{\tiny$\pm$0.53} (1.36$\uparrow$) & 61.93{\tiny$\pm$0.36} (1.11$\uparrow$) & 67.01{\tiny$\pm$0.77} (0.08$\uparrow$) & 67.40{\tiny$\pm$0.38} (1.16$\uparrow$) \\
        + GradDrop \cite{DBLP:conf/accv/TsengCTLL020}& 45.57{\tiny$\pm$0.37} (1.16$\uparrow$) & 52.30{\tiny$\pm$0.47} (0.14$\uparrow$)   & 54.54{\tiny$\pm$0.50} (2.71$\uparrow$) & 62.35{\tiny$\pm$0.38} (1.53$\uparrow$) & 67.33{\tiny$\pm$0.38} (0.40$\uparrow$) & 67.03{\tiny$\pm$0.42} (0.79$\uparrow$) \\
        + MetaMix \cite{DBLP:conf/icml/YaoHZ0TZHL21} & 46.06{\tiny$\pm$0.44} (1.65$\uparrow$) & 53.32{\tiny$\pm$0.49} (1.16$\uparrow$)   & 54.92{\tiny$\pm$0.38} (3.09$\uparrow$) & \textbf{62.86{\tiny$\pm$0.38}} (2.04$\uparrow$) & 68.81{\tiny$\pm$0.41} (1.88$\uparrow$) & 68.37{\tiny$\pm$0.36} (2.13$\uparrow$) \\
        \midrule
        + \textbf{MGAug-WP}         & 45.65{\tiny$\pm$0.32} (1.24$\uparrow$) & 52.97{\tiny$\pm$0.45} (0.81$\uparrow$)   & 54.67{\tiny$\pm$0.42} (2.84$\uparrow$) & 61.34{\tiny$\pm$0.40} (0.52$\uparrow$) & 67.53{\tiny$\pm$0.39} (0.60$\uparrow$) & 67.75{\tiny$\pm$0.47} (1.51$\uparrow$)  \\
        + \textbf{MGAug-PP}         & 45.77{\tiny$\pm$0.30} (1.36$\uparrow$) & \underline{54.53{\tiny$\pm$0.45}} (2.37$\uparrow$)   & 54.60{\tiny$\pm$0.40} (2.77$\uparrow$) & 62.10{\tiny$\pm$0.41} (1.28$\uparrow$) & 68.95{\tiny$\pm$0.38} (2.02$\uparrow$) & \underline{68.41{\tiny$\pm$0.44}} (2.17$\uparrow$)  \\
        + \textbf{MGAug-CP}         & 45.95{\tiny$\pm$0.41} (1.54$\uparrow$) & \textbf{54.70{\tiny$\pm$0.46}} (2.54$\uparrow$)   & \textbf{56.24{\tiny$\pm$0.40}} (4.41$\uparrow$) & \underline{62.41{\tiny$\pm$0.40}} (1.59$\uparrow$) & \textbf{69.27{\tiny$\pm$0.38}} (2.34$\uparrow$) & \textbf{68.69{\tiny$\pm$0.44}} (2.45$\uparrow$)  \\
        \midrule
        + \textbf{MGAug-MaxUp-WP}   & 45.78{\tiny$\pm$0.28} (1.37$\uparrow$) & 53.41{\tiny$\pm$0.39} (1.25$\uparrow$)   & 54.73{\tiny$\pm$0.44} (2.90$\uparrow$) & 61.23{\tiny$\pm$0.38} (0.41$\uparrow$) & 68.60{\tiny$\pm$0.38} (1.67$\uparrow$) & 67.99{\tiny$\pm$0.50} (1.75$\uparrow$)  \\
        + \textbf{MGAug-MaxUp-PP}   & \underline{46.30{\tiny$\pm$0.30}} (1.89$\uparrow$) & 53.83{\tiny$\pm$0.39} (1.67$\uparrow$)   & 55.24{\tiny$\pm$0.47} (3.41$\uparrow$) & 62.03{\tiny$\pm$0.38} (1.21$\uparrow$) & \underline{69.04{\tiny$\pm$0.40}} (2.11$\uparrow$) & 67.90{\tiny$\pm$0.53} (1.66$\uparrow$)  \\
        + \textbf{MGAug-MaxUp-CP}   & \textbf{46.42{\tiny$\pm$0.37}} (2.01$\uparrow$) & 54.15{\tiny$\pm$0.37} (1.99$\uparrow$)   & \underline{55.35{\tiny$\pm$0.39}} (3.52$\uparrow$) & 62.24{\tiny$\pm$0.41} (1.42$\uparrow$) & 68.65{\tiny$\pm$0.38} (1.72$\uparrow$) & 68.10{\tiny$\pm$0.47} (1.86$\uparrow$)  \\
        \bottomrule
    \end{tabular}
\end{table*}

\subsection{Experimental Settings} \label{sec:exp_setting}

\subsubsection{Datasets}
We conduct experiments with two widely-used datasets: \emph{mini}-ImageNet ~\cite{DBLP:conf/nips/VinyalsBLKW16} and CUB ~\cite{wah2011caltech}. \emph{Mini}-ImageNet consists of $100$ classes of natural images sampled from ImageNet ~\cite{DBLP:conf/cvpr/DengDSLL009}, with $600$ images per class. It is split into non-overlapping $64$, $16$, and $20$ classes for training, validation, and testing.
The CUB contains $200$ species of birds and $11,788$ images in total. We randomly select $100$ classes as the training set, and the others are equally divided for validation and testing. 
Experiments involve $5$-way $1$-shot and $5$-shot tasks in both mutually exclusive (ME) and non-mutually-exclusive (NME) settings following the previous work \cite{DBLP:conf/iclr/YinTZLF20}, where each task contains $15$ query samples.

\subsubsection{Backbones}
We use two backbones with different depths, including Conv-$4$, ResNet-$10$, and ResNet-$18$. The Conv-$4$ contains four convolution blocks, each block is concatenated by convolution, BatchNorm, nonlinear activation (ReLU), and max pooling layers. The ResNet-$10$ is a simplified ResNet-$18$ ~\cite{DBLP:conf/cvpr/HeZRS16} where only one residual building block is used in each layer. Following previous works ~\cite{DBLP:conf/iclr/ChenLKWH19,DBLP:conf/nips/YangZ020}, we respectively resize images to $84 \times 84$ and $224 \times 224$ before feeding the Conv and ResNet backbones and correspondingly randomly scale to [$84$, $64$, $48$] and [$224$, $192$, $160$, $140$] as the data augmentation of sub-networks.

\subsubsection{Baselines}
We choose MAML ~\cite{DBLP:conf/icml/FinnAL17} and Prototypical Network ~\cite{DBLP:conf/nips/SnellSZ17} (abbreviated as ProtoNet) as instance baselines. The former belongs to the GBML branch, and the latter is a classic MBML method. For MAML, we implement a first-order approximation FoMAML for efficiency ~\cite{DBLP:conf/icml/FinnAL17}. We further take the transformations designed in Baseline++ ~\cite{DBLP:conf/iclr/ChenLKWH19} as the data regularization baseline and mark it with `Aug'. In the following experiments, we mark the pruning strategy with ``-XX'' and make MGAug-CP as the default setting, abbreviated as MGAug.

\subsubsection{Implementation details}
For $1$-shot tasks, we respectively train $4800$ and $1600$ epochs for GBML and MBML methods, and each epoch includes $100$ episodes. For $5$-shot tasks, the number of epochs is halved. All results are average results over $600$ episodes with confidence intervals of radius one standard error. Following the training procedure of ~\cite{DBLP:conf/iclr/ChenLKWH19}, all methods are trained from scratch and use the Adam optimizer with an initial learning rate of $10^{-3}$.


\subsection{Comparison with existing meta-regularization strategies} \label{sec:exp_sota}
\subsubsection{MBML-based strategies}
Table \ref{tab:result_proto_cub} and \ref{tab:result_proto_mini} list the results of ProtoNet baseline with different meta-regularization methods on CUB and \emph{mini}-ImageNet, respectively, where the best results are marked in bold and the second with an underline. In addition to Aug ~\cite{DBLP:conf/iclr/ChenLKWH19}, we also compare two state-of-the-art regularization methods designed for the MBML branch, including TaskAug \cite{DBLP:journals/corr/abs-2003-00804}, Meta-MaxUp \cite{DBLP:conf/icml/NiGSKG21}. Results show that memorization breaking and augmented diversity gradients greatly improve classification accuracy. For example, in the $5$-way $1$-shot + ResNet-$10$ scenario, MGAug improves the accuracy by $6.96\%$ and $6.67\%$ on CUB and \emph{mini}-ImageNet, respectively.

\subsubsection{GBML-based strategies} \label{sec:Exp_GBML}
Table \ref{tab:result_fomaml_cub} and \ref{tab:result_fomaml_mini} list results of FoMAML baseline with different meta-regularization methods on CUB and \emph{mini}-ImageNet, respectively, where the best results are marked in bold and the second with an underline. We compare four state-of-the-art regularization methods designed for the GBML branch, including MR ~\cite{DBLP:conf/iclr/YinTZLF20}, TAML ~\cite{DBLP:conf/cvpr/JamalQ19}, MetaMix ~\cite{DBLP:conf/icml/YaoHZ0TZHL21} and GradDrop ~\cite{DBLP:conf/accv/TsengCTLL020}. 
The former two design explicit regularization terms to address memorization and task bias issues in fast adaptation, respectively. While the latter two are typical methods of data and gradient regularization, where MetaMix mixes the input and its features using the MixUp strategy and GradDrop randomly drops meta-gradients to increase its diversity. Compared to random-based strategies, gradient diversity in MGAug is learned by different sub-networks on the same task, which leads to self-guided augmentation and higher classification accuracy.

\subsubsection{Loss and accuracy curves}
Besides accuracy, we plot the loss and accuracy curves in Fig. \ref{fig:loss} to observe meta-overfitting in FoMAML baseline, Aug, and MGAug. Among them, the FoMAML baseline has the lowest loss and the highest accuracy during training, especially in CUB, while it performs the worst over validation epochs. This inversion is powerful evidence of overfitting. In contrast, Aug and MGAug do not significantly overfit the training task and generalize better on unseen tasks. Another interesting trend is the trade-off between training loss and validation accuracy. The training loss of MGAug is always lower than Aug, but it yields more accurate predictions. This phenomenon means that MGAug learns more generalizable meta-knowledge during training. In other words, even well-designed data transformations may potentially inhibit the representation capability of the network.

\begin{figure}[t]
    \centering
    \includegraphics[width=0.96\linewidth]{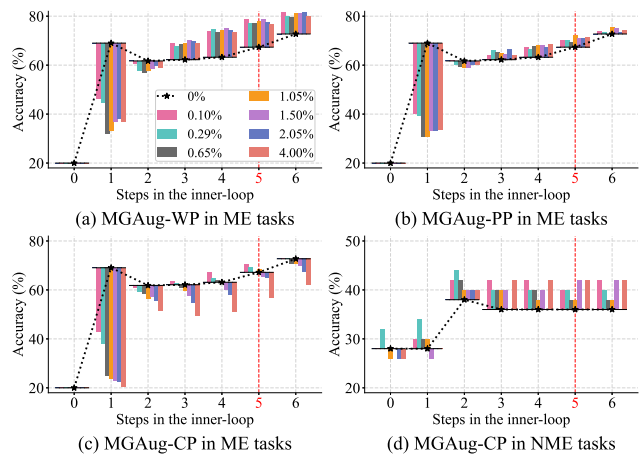}
    \caption{Hat graphs of the base learner's accuracy on ME tasks with (a) WP, (b) PP, and (c) CP. As a comparison to (c), (d) shows the result of MGAug-CP on NME tasks. The dotted line indicates the results of a full network (i.e., 0\% pruning), and the histogram indicates the gap between the accuracy of full and sub-networks with different pruning rates.}
    \label{fig:break}
\end{figure}

\subsection{Behavioral Analysis of MGAug} \label{sec:exp_analysis}
\subsubsection{Rote memorization breaking}
\begin{figure}[t]
        \centering
        \includegraphics[width=0.6\linewidth]{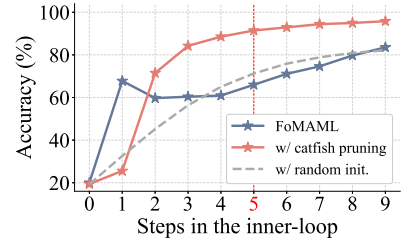}
        \caption{Accuracy comparison of the base learner after fine-tuning with and without \textit{catfish pruning}.}
        \label{fig:break_acc}
\end{figure}

We observe the behavior of the base learner to investigate the memorization breaking in the inner loop. To this end, we visualize the gap of fine-tuning accuracy between the full-network and sub-networks with different pruning rates via a modified hat graph ~\cite{witt2019introducing}. Fig. \ref{fig:break} shows the average accuracy of FoMAML with MGAug using ResNet-$10$ on $100$ tasks sampled from CUB. Results for the full-network are indicated by the dashed line with asterisks. The histogram reflects the gap between the accuracy of sub-networks and the full-network, i.e., the upward bar indicates higher accuracy than the full-network and vice versa. The trend in Fig. \ref{fig:break} reflects whether fine-tuning relies on rote memorization or rapid adaptation of meta-knowledge. 
\begin{enumerate}[i.]
    \item \textbf{NME tasks suffer from severe memorization overfitting.} Observing Fig. \ref{fig:break} (c) and (d), due to label mutual exclusion, each ME task cannot be handled solely on memorization, i.e., the accuracy at step-$0$ is similar to random classification and improves rapidly after fine-tuning. In contrast, for NME tasks, there is rote memorization about training samples in meta-learned parameters that resulted in $28$\% classification accuracy without fine-tuning (at step-$0$) and further limited the fine-tuning performance.
    
    \item \textbf{The ME setting is short-lived for solving memorization issues.} Although the ME setting avoids the reliance on memorization at step-$0$, the accuracy increases sharply after only one step and remains almost unchanged until step $5$ (fine-tuning five steps by default ~\cite{DBLP:conf/icml/FinnAL17}). This trend means that memorization is almost recovered with just one iteration and still prevents subsequent fine-tuning.
    
    \item \textbf{Our MGAug breaks rote memorization.} Unlike constructing ME tasks, MGAug directly breaks memorization and inhibits its recovery. An intuitive phenomenon is that accuracy slowly increases during fine-tuning, even with only a $0.1\%$ pruning rate. Following the same ME setting, Fig. \ref{fig:break_acc} visualizes fine-tuning curves to verify whether fine-tuning is reactivated with broken memorization, which is the key to overcoming memorization overfitting. Further, we plotted the curve fine-tuned from random initialization as a baseline with no memory at all. Clearly, the trend of MGAug is closer to random initialization than to constructing ME tasks, indicating that the memorization issue is significantly alleviated and fine-tuning is reactivated. Also, the accuracy of MGAug improves more quickly than random initialization, implying a faster adaptation of meta-knowledge to new tasks.

    \item \textbf{CP has stronger breaking capability than WP and PP.} Although all three effectively hinder memory recovery, CP is clearly the most effective, followed by PP and finally WP, as seen in step-$1$ in Fig. \ref{fig:break} (a), (b), and (c).

\end{enumerate}

\begin{figure}[t]
    \centering
    \includegraphics[width=0.6\linewidth]{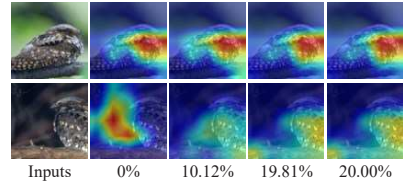}
    \caption{Grad-CAM visualization of two representative samples in CUB.}
    \label{fig:break_cam}
\end{figure}

\subsubsection{Augmented meta-gradients}
We empirically infer that the effectiveness of the meta-gradient augmentation derived from pruned sub-networks is twofold. One is the improvement from resolving memorization overfitting, which has been verified in the previous section. The other is the diversity of attention introduced by sub-networks with different pruning rates, even for the same task. To verify this, Fig. \ref{fig:break_cam} visualizes the attention regions of different sub-networks using Grad-CAM ~\cite{DBLP:journals/ijcv/SelvarajuCDVPB20} and lists representative examples. Interestingly, attention changes seem to occur more often in samples containing insignificant objects (bottom). Conversely, for the salient ones (top), the learner is more confident in the predictions.


\subsubsection{Plug-and-play property}
\begin{figure*}
\centering
  \includegraphics[width=0.93\linewidth]{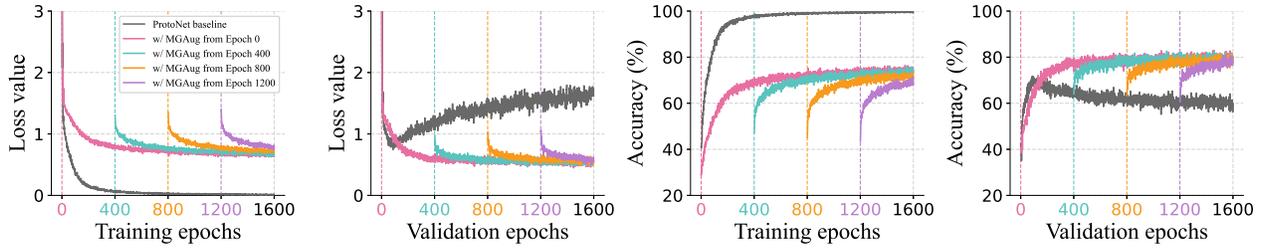}
  \caption{Loss and accuracy curves of ProtoNet integrating MGAug at different epochs. The first two are loss curves for training and validation, and the last two are the corresponding accuracy trends.}\label{fig:warmup}
\end{figure*}

In addition, MGAug can also improve meta-generalization in a flexible plug-and-play way. Fig. \ref{fig:warmup} shows the results of training with MGAug starting from epochs $0$, $400$, $800$, and $1200$ on $5$-way $5$-shot CUB tasks. Both train and test curves show that MGAug consistently improves ProtoNet baseline performance and avoids meta-overfitting, even if it is only used for the last $400$ epochs.


\subsubsection{The comparison of training costs} 
The additional overhead required by MGAug is related to the number $U$ of sub-networks, more precisely $U$ times the cost of the vanilla model. Although it can be accelerated by parallel computing the sub-network, we still designed the lightweight MGAug-MaxUp to trade off performance and overhead. Table \ref{tab:time} lists the time for meta-training once on a single $5$-way $5$-shot task under the same environment. It can be seen that MGAug-MaxUp performs similarly to the vanilla methods with almost no additional computational cost.

\begin{table}[!t]
    \centering
    \footnotesize
    \setlength\tabcolsep{4pt}
    \caption{Average cost of meta-training once with three sub-networks per $5$-way $5$-shot CUB task (seconds). \label{tab:time}}
    \begin{tabular}{l c c c c}
        \toprule
         & Conv-$4$  & ResNet-$10$  &  ResNet-$18$\\
        \midrule
        FoMAML    & $0.136$    & $0.279$  & $0.510$    \\
        + \textit{MGAug}        & $0.407$    & $0.942$ & $1.702$ \\
        + \textit{MGAug-MaxUp}  & $0.135$   & $0.286$  & $0.473$ \\
        \midrule
        ProtoNet    & $0.029$    & $0.078$  & $0.144$   \\
        + \textit{MGAug}        & $0.089$    & $0.315$ & $0.575$  \\
        + \textit{MGAug-MaxUp}  & $0.031$   & $0.087$  & $0.145$ \\
        \bottomrule
    \end{tabular}   
\end{table}

\begin{figure}[!t]
\centering
  \includegraphics[width=0.88\linewidth]{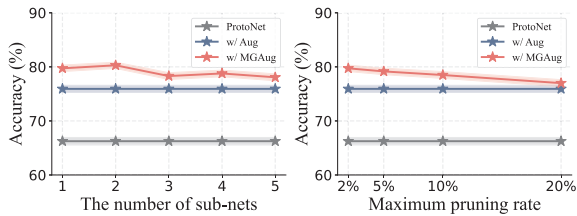}
  \caption{Comparison of accuracy for different numbers of sub-networks (left) and range of pruning rates (right) based on the ProtoNet baseline.}\label{appendix-fig:alb}
\end{figure}

\begin{table*}[!t]
    \centering
    \footnotesize
    \caption{Classification results of more meta-baselines on $5$-way $1$-shot CUB tasks. \label{tab:extend_baseline}}
    \begin{tabular}{l l l l l}
        \toprule
                    & Baseline        &  + Aug         & + \textit{MGAug}      & + \textit{MGAug-MaxUp} \\
         \midrule
         Reptile \cite{DBLP:journals/corr/abs-1803-02999}   & 51.48{\tiny$\pm$0.45} & 50.16{\tiny$\pm$0.44} & 54.62{\tiny$\pm$0.46} (4.46$\uparrow$) & \textbf{55.26{\tiny$\pm$0.45}} (5.10$\uparrow$)      \\
         CAVIA \cite{DBLP:conf/icml/ZintgrafSKHW19}         & 53.67{\tiny$\pm$0.99} & 54.11{\tiny$\pm$0.99} & 58.28{\tiny$\pm$1.04} (4.17$\uparrow$) & \textbf{58.31{\tiny$\pm$1.05}} (4.20$\uparrow$)    \\
         MAML  \cite{DBLP:conf/icml/FinnAL17}               & 55.16{\tiny$\pm$0.48} & 60.25{\tiny$\pm$0.49} & \textbf{61.03{\tiny$\pm$0.48}} (0.78$\uparrow$) & 60.58{\tiny$\pm$0.49} (0.33$\uparrow$)   \\
         \midrule
         R$2$-D$2$ \cite{DBLP:conf/iclr/BertinettoHTV19}      & 56.27{\tiny$\pm$0.97} & 61.66{\tiny$\pm$0.94} & 60.89{\tiny$\pm$0.94} (0.77$\downarrow$)          & \textbf{62.62{\tiny$\pm$0.95}} (0.96$\uparrow$)     \\
         MetaOptNet \cite{DBLP:conf/cvpr/LeeMRS19}  & 57.60{\tiny$\pm$0.94} & 57.34{\tiny$\pm$0.91} & \textbf{60.62{\tiny$\pm$0.94}} (3.28$\uparrow$)    & 59.51{\tiny$\pm$0.95} (2.17$\uparrow$)      \\
          \bottomrule
    \end{tabular}
\end{table*}

\subsection{Robustness experiments on hyper-parameters}  \label{sec:exp_hyperparams}
This subsection aims to evaluate the robustness of MGAug to hyper-parameters. Specifically, the hyper-parameters mainly involve the number of sub-networks and the range of pruning rates. The following experiments are all performed on the $5$-way $5$-shot CUB tasks using the Conv-$4$ backbone.

\subsubsection{The number of sub-networks} Fig. \ref{appendix-fig:alb} (left) shows the results of ProtoNet with MGAug using different numbers of sub-networks (from one to five). Clearly, MGAug outperforms the baseline for all hyper-parameter settings. Even pruning one sub-network for each task still improves classification accuracy, which validates the theoretical analysis in Theorem \ref{theorem1}. In other experiments, three and one sub-network empirically default to the GBML and MBML branches, respectively.

\begin{table}[!t]
\centering
\footnotesize
\setlength\tabcolsep{4pt}
\caption{$1$-shot accuracy under the cross-domain scenario with a Conv-$4$ backbone.\label{tab:extend_cross}}
    \begin{tabular}{l l l}
        \toprule
        & \emph{mini} $\to$ CUB & CUB $\to$ \emph{mini}\\
        \midrule
        FoMAML                 & 36.26{\tiny$\pm$0.33}         & 28.92{\tiny$\pm$0.28}        \\       
        + Aug                  & 33.93{\tiny$\pm$0.30}         & 29.02{\tiny$\pm$0.32}        \\       
        + \textit{MGAug}       & \textbf{37.69{\tiny$\pm$0.35}} (3.76$\uparrow$)   & \textbf{31.96{\tiny$\pm$0.34}} (2.94$\uparrow$)  \\
        + \textit{MGAug-MaxUp} & 34.55{\tiny$\pm$0.32} (0.62$\uparrow$)   & 29.58{\tiny$\pm$0.31} (0.56$\uparrow$) \\
        \midrule
        ProtoNet               & 29.07{\tiny$\pm$0.59}         & 26.55{\tiny$\pm$0.49}        \\     
        + Aug                  & 32.49{\tiny$\pm$0.63}         & 31.90{\tiny$\pm$0.67}        \\       
        + \textit{MGAug}       & \textbf{35.98{\tiny$\pm$0.68}} (3.49$\uparrow$)   &  32.35{\tiny$\pm$0.65} (0.45$\uparrow$)  \\        
        + \textit{MGAug-MaxUp} & 34.89{\tiny$\pm$0.66} (2.40$\uparrow$)   & \textbf{34.60{\tiny$\pm$0.71}} (2.70$\uparrow$)  \\
        \bottomrule
    \end{tabular}
\end{table}

\begin{table}[!t]
    \centering
    \footnotesize
    \setlength\tabcolsep{4pt}
    \caption{Classification results of ProtoNet using deeper backbones on $5$-way $1$-shot CUB tasks. \label{tab:extend_backbone}}
    \begin{tabular}{l l l}
        \toprule
         & ResNet-$34$  & ResNet-$50$\\
        \midrule
        ProtoNet                & 58.53{\tiny$\pm$1.04} & 61.06{\tiny$\pm$1.04}      \\       
        + Aug                   & 74.21{\tiny$\pm$0.95} & 73.87{\tiny$\pm$0.95}      \\       
        + \textit{MGAug}        & 76.10{\tiny$\pm$0.96} (1.89$\uparrow$)  & \textbf{79.72{\tiny$\pm$0.96}} (5.85$\uparrow$) \\
        + \textit{MGAug-MaxUp}  & \textbf{77.61{\tiny$\pm$0.99}} (3.40$\uparrow$) & 78.61{\tiny$\pm$0.99} (4.74$\uparrow$) \\
        \bottomrule
    \end{tabular}   
\end{table}

\subsubsection{The range of pruning rates}  Fig. \ref{appendix-fig:alb} (right) shows the results of four range pruning rates, where the largest pruning rates are $2\%$, $5\%$, $10\%$ and $20\%$, respectively. As the pruning rate increases, the accuracy gradually decreases. This is because the proposed memorization-breaking criterion preferentially removes parameters carrying meta-memorization. When the memorization is completely erased, meta-learning degenerates into ordinary few-shot learning. Nevertheless, even with the $20\%$ setting, MGAug still improves the baseline, indicating its robustness to hyper-parameters.

\subsection{Robustness experiments on other scenarios} \label{sec:exp_other}
To verify the flexibility and generality of MGAug, we provide experimental results in more scenarios, including more meta-baselines, backbones, and tasks. The experiments in this subsection are based on $5$-way $1$-shot tasks.

\subsubsection{More meta-learning instances} We first supplement the results of FoMAML and ProtoNet with a Conv-$4$ backbone on the \emph{mini}-ImageNet dataset in Table \ref{tab:extend_baseline}, and then integrate MGAug into more meta-baselines, including Reptile ~\cite{DBLP:journals/corr/abs-1803-02999}, CAVIA ~\cite{DBLP:conf/icml/ZintgrafSKHW19}, MAML ~\cite{DBLP:conf/icml/FinnAL17}, R$2$-D$2$ ~\cite{DBLP:conf/iclr/BertinettoHTV19}, and MetaOptNet ~\cite{DBLP:conf/cvpr/LeeMRS19}. The first three are instances of the GBML, and the latter two belong to a branch called \emph{last-layer meta-learning} ~\cite{DBLP:conf/icml/GoldblumRFNCG20}. The classification results listed in Table \ref{tab:extend_baseline} show that MGAug significantly improves the performance of these meta-learning methods compared to the data augmentation (i.e., Aug), especially in Reptile, CAVIA, and MetaOptNet. 

\subsubsection{Deeper backbones} Since the underlying assumption is that deep models are more susceptible to overfitting, we explore the performance of MGAug on the deeper ResNet-$34$ and ResNet-$50$ backbone ~\cite{DBLP:conf/cvpr/HeZRS16}. Table \ref{tab:extend_backbone} lists the classification accuracy of the MGAug based on the ProtoNet baseline. Compared with shallower backbones, data augmentation improves accuracy significantly in deeper ones, while MGAug can further gain about $3$\% improvement.


\subsubsection{Cross-domain tasks} To further evaluate how MGAug improves generalization of meta-learning methods, we conduct a cross-domain experiment in which the test set is from an unseen domain. Following the cross-domain setting in ~\cite{DBLP:conf/iclr/ChenLKWH19,DBLP:conf/accv/TsengCTLL020}, meta-learner is trained on \emph{mini}-ImageNet and evaluated on the few-shot task constructed on CUB, or vice versa. The classification accuracy on the $5$-way $1$-shot task is shown in Table \ref{tab:extend_cross}, where the models trained on \emph{mini}-ImageNet perform better overall than those trained on CUB, benefiting from the rich categories of training samples. In both cross-domain settings, our method still significantly improves accuracy, indicating that MGAug induces the model to meta-learn more transferable features.

\section{Conclusion}
This work proposes a data-independent meta-regularization method, termed MGAug, which can alleviate both memorization and learner overfitting in the two-loop meta-learning framework. Unlike existing task augmentation and explicit regularization terms, the key idea is to first solve the rote memorization issue and restore adaptability in the inner loop via network pruning, and then alleviate learner overfitting with augmented meta-gradients derived from pruned sub-networks. We explore two random pruning strategies and propose a noval \emph{catfish pruning} that achieves the most significant memorization breaking by removing the parameters containing the largest amount of rote memories. We also deduce a PAC-Bayes-based generalization bound for MGAug and further implement a lightweight version balancing performance and overhead. Extensive experimental results show that MGAug significantly outperforms existing meta-learning baselines. Meanwhile, we believe that MGAug's ideas and implementations can also inspire and drive the development of gradient regularization strategies.

{\appendices
\section*{Appendix}

\subsection{Proof of Proposition 1} \label{appendix1}
For the $t$-th task, Meta-Memorization Carrying Amount (MMCA) is defined as the difference in query loss before and after pruning parameter $\theta_{(j)}$, i.e.,
\begin{equation}
\begin{footnotesize}
    \begin{aligned}
        \text{MMCA}_{t, (j)} &\triangleq \Delta \mathcal{L}_{(j)}(\theta; D^q_t) \\ 
        & = \mathcal{L}\left(\mathbf{1} \odot \theta; D^q_t\right) - \mathcal{L}\left(\left(\mathbf{1}-e_{(j)}\right) \odot \theta; D^q_t\right),
    \end{aligned}
    \label{Appendix-eq:10_catfish_MMCA}
\end{footnotesize}
\end{equation} 
where $\mathbf{1}$ is the vector of dimension $n$ and $\odot$ is the Hadamard product. The $e_{(j)}$ is the indicator vector for the $j$-th parameter $\theta_{(j)}$, whose value is zero everywhere except that index $j=1$. 

In essence, the calculation of $\text{MMCA}_{t,(j)}$ is to measure the the effect of the $j$-th initial parameter on the loss function. We additionally introduce a pruning indicator variable $c\in \{0,1\}^n$, where $c_{(j)}$ indicates whether the parameter $\theta_{(j)}$ is preserved ($c_{(j)}=1$) or pruned ($c_{(j)}=0$). Further, the optimization objective of the base learner in inner loop can be rewritten as $\mathop{\arg\min}_{\theta} \mathcal{L}(c \odot \theta(\omega), D^s_t)$. Obviously, $\Delta \mathcal{L}_{(j)}(\theta; D^q_t)$ can be approximated as the derivative of $\mathcal{L}$ with respect to $c_{(j)}$. But, since $c$ is binary, $\mathcal{L}$ is not differentiable with respect to $c$ in this discrete setting. Therefore, by relaxing the binary constraint in the indicator variable $c$, the effect of parameter $\theta_{(j)}$ on the loss can be approximated as: 
\begin{equation}
\begin{footnotesize}
    \begin{aligned}
        \Delta \mathcal{L}_{(j)}(\theta; D^q_t) & \approx \left.\frac{\partial \mathcal{L}(c \odot \theta, D^q_t) }{\partial c_{(j)}} \right|_{c=1} \\ 
        & = \lim_{\delta \to 0} \left.\frac{\mathcal{L}\left(c \odot \theta, D^q_t\right) - \mathcal{L}\left((c-\delta e_{(j)}) \odot \theta, D^q_t\right)}{\delta} \right|_{c=1}.
    \end{aligned}
    \label{Appendix-eq:11_limits}
\end{footnotesize}
\end{equation}

Clearly, $\partial \mathcal{L}/ \partial c_{(j)}$ is an infinitesimal version of $\Delta \mathcal{L}_{(j)}$, that measures the rate of change of $\mathcal{L}$ with respect to an infinitesimal change in $c_{(j)}$ from $1 \to 1 - \delta$. This can be computed efficiently in one forward-backward pass using automatic differentiation, for all $j$ at once ~\cite{DBLP:conf/iclr/LeeAT19}.

Assumes $c_{(j)} = 1$. Let $a_{(j)}$ be the incoming activation that is multiplied by $\theta_{(j)}$, and $z$ be the pre-activation of the neuron to which $\theta_{(j)}$ serves as an input, i.e., $z = c_{(j)} a_{(j)} \theta_{(j)}$. According to the given conditions and the chain rule, we can deduce the MMCA value by 
\begin{equation}
\begin{footnotesize}
    \begin{aligned}
        \text{MMCA}_{t, (j)} \approx \frac{\partial \mathcal{L}}{\partial c_{(j)}} & = \frac{\partial \mathcal{L}}{\partial z} \frac{\partial z}{\partial c_{(j)}} = \frac{\partial \mathcal{L}}{\partial z} a_{(j)} \theta_{(j)} \\
        & = \frac{\partial \mathcal{L}}{\partial z} \frac{\partial z}{\partial \theta_{(j)}} \theta_{(j)} = \frac{\partial \mathcal{L}}{\partial \theta_{(j)}} \theta_{(j)}.
    \end{aligned}
    \label{eq:5_catfish_MMCA}
\end{footnotesize}
\end{equation}

Therefore, the MMCA score of a parameter is essentially determined by its weights and derivatives. The weights represent the state of the meta memorization (knowledge), while the derivatives represent which memorization is sensitive to the current task. By removing the parameter with a large MMCA score, rote memorization is efficiently broken.

\subsection{Proof of Theorem 1} \label{appendix2}
This section provides proof of Theorem 1. Following the previous work \cite{DBLP:conf/icml/AmitM18}, the proof begins with McAllaster’s classical PAC-Bayes bound \cite{DBLP:conf/colt/McAllester99} for a single task and consists of two steps. In the first step, we bound the errors caused by observing insufficient samples in each task, and each task is assigned a pruned sub-network. In the second step, we bound the error caused by observing a limited number of tasks in the environment. 

\begin{theorem} \textbf{\emph{(McAllester's single-task bound \cite{DBLP:conf/colt/McAllester99}).}} Let $\mathcal{X}$ be a sample space and $\mathbb{X}$ some distribution over $\mathcal{X}$, and let $\mathcal{F}$ be a hypothesis space of functions over $\mathcal{F}$. Define a `loss function' $g(f, X):\mathcal{F}\times\mathcal{X}\to[0, 1]$, and let $X_1^M:={X_1,\dots,X_M}$ be a sequence of $M$ independent random variables distributed according to $\mathbb{X}$. Let $\pi$ be some prior distribution over $\mathcal{F}$ (which must not depend on the samples $X_1$,\dots,$X_M$). For any $\delta \in (0, 1]$, the following bound holds uniformly for all ‘posterior’ distributions $\kappa$ over $\mathcal{F}$ (even sample-dependent),

\begin{equation}
\begin{footnotesize}  
    \begin{aligned}
        \mathbb{P}_{X_1^M\underset{i.i.d}{\sim}\mathbb{X}}&\left\{\underset{X\sim\mathbb{X}}{\mathbb{E}} \underset{f\sim\kappa}{\mathbb{E}} g(f,X)\leq\frac{1}{M}\sum_{m=1}^M \underset{f\sim\kappa}{\mathbb{E}}g(f,X_m) \right. \\ 
        & \left. + \sqrt{\frac{1}{2(M-1)}\left(D(\kappa\Vert\pi)+\log\frac{M}{\delta}\right)},\forall \kappa \right\} \ge 1-\delta.
    \end{aligned}
    \label{eq:appendix_bound_4}
\end{footnotesize}
\end{equation}
\end{theorem}

\subsubsection{First step} We use Theorem 2 to bound the generalization error in each of the observed tasks with a meta-learned algorithm $\mathcal{Q}$. 
Let $i\in 1,\dots, T$ be the index of observed tasks. The samples are $X_m:=z_{i,j}$, the number of samples is $M:=m_i$, and sample distribution is $\mathbb{X}:=\mathcal{D}_i$. The `loss function' is $g(f, X):=l(h,z)$. We define the `prior over hypothesis' $\pi:=(\mathcal{P}, P)$, in which we first sample $P$ from $\mathcal{P}$ and then sample hypothesis $h$ from $P$. According to Theorem 2, the `posterior over hypothesis' can be any distribution, in particular, the bound will hold for the following family of distributions $\kappa:=(\mathcal{Q}, Q)$, where we first sample $P$ from $\mathcal{Q}$ and then sample $h$ from $Q = Q(\mathcal{T}_i, P)$ with the task $\mathcal{T}_i$. For deep-network-based methods, the meta-learned algorithm typically refers to the meta-parameters $\Theta \in \mathbb{R}^d$. Given a dropout rate $\rho\in [0, 1]$, we can generate sub-network parameters $\theta \in \mathbb{R}^d$ by pruning based on the proposed criterion. For each coordinate $\theta^i$, the value 0 with probability $\rho$ (pruning the coordinate $\theta^i$) or with probability $1-\rho$ setting $\theta^i = \Theta^i + \epsilon$, where $\epsilon \sim \mathcal{N}(0, 1)$ is an auxiliary noise vector. Let $Q_{\rho, \Theta}$ denote the distribution on parameter vectors defined by this pruning process, and the `prior' and `posterior' distributions can be re-marked as $Q_{\rho, 0}$ and $Q_{\rho, \Theta}$. 

To further clarify the formal notation of the pruning process, we consider the Boolean d-cube $\mathcal{B}$ which is the set of pruning mask vector $s\in\mathbb{R}^d$ such that $s_i\in\{0,1\}$ for all $1\leq i\leq d$. Following the previous work \cite{DBLP:journals/corr/McAllester13}, $s\in \mathcal{B}$ is called the ``sparsity patterns''. We let $S_\rho$ be the distribution on the sparsity patterns generated by selecting each $s^i$ independently with the probability of $s^i=0$ being $\rho$. For a given  $s$ and $\theta\in \mathbb{R}^d$ we will write $s\circ \theta$ for the Hadamard product defined by $(s\circ \theta)^i=s^i\theta^i$. We then have that a draw from $Q_{\rho, \Theta}$ can be made by first drawing a sparsity pattern $s\sim S_{\rho}$ and a noise vector $\epsilon \sim \mathcal{N}(0, 1)$, and then constructing product defined by $s\circ (\Theta + \epsilon)$, i.e., $\underset{\theta \sim Q_{\rho, \Theta}}{\mathbb{E}}(f(\theta)) = \underset{s \sim S_{\rho}}{\mathbb{E}}\left(f(s\circ (\Theta + \epsilon))\right)$. 

The KL-divergence term is 
\begin{equation}
\begin{footnotesize}  
    \begin{aligned}
    D(\kappa\Vert \pi) &= \underset{f\sim \kappa}{\mathbb{E}} \log \frac{\kappa(f)}{\pi(f)} = \underset{Q_{\rho, 0}\sim \mathcal{Q}}{\mathbb{E}} \underset{h\sim Q_{\rho, \Theta}}{\mathbb{E}} \log \frac{\mathcal{Q}(Q_{\rho, 0})Q_{\rho, \Theta}(h)}{\mathcal{P}(Q_{\rho, 0})Q_{\rho, 0}(h)}\\
    &= \underset{P\sim \mathcal{Q}}{\mathbb{E}} \log \frac{\mathcal{Q}(P)}{P(P)} + \underset{Q_{\rho, 0}\sim \mathcal{Q}}{\mathbb{E}} \underset{h\sim Q_{\rho, \Theta}}{\mathbb{E}} \log \frac{Q_{\rho, \Theta}(h)}{Q_{\rho, 0}(h)}\\
    &= D(\mathcal{Q}\Vert \mathcal{P}) + \underset{Q_{\rho, 0}\sim \mathcal{Q}}{\mathbb{E}} D(Q_{\rho, \Theta}\Vert Q_{\rho, 0})\\
    &= D(\mathcal{Q}\Vert \mathcal{P}) + \underset{s\sim S_\rho}{\mathbb{E}} \underset{\epsilon \sim \mathcal{N}(0, 1)}{\mathbb{E}} \ln\frac{S_\rho(s)e^{-\frac{1}{2}\Vert s\circ \epsilon\Vert^2}}{S_{\rho}(s)e^{-\frac{1}{2}\Vert s\circ (\Theta+\epsilon) \Vert^2}}\\
    &= D(\mathcal{Q}\Vert \mathcal{P}) +  \underset{s\sim S_\rho}{\mathbb{E}} \left(\frac{1}{2}\Vert s\circ \Theta \Vert^2 \right)\\
    &= D(\mathcal{Q}\Vert \mathcal{P}) +  \frac{1-\rho}{2}\Vert \Theta \Vert^2
    \label{eq:appendix_bound_5}
    \end{aligned}
\end{footnotesize}
\end{equation}

Plugging in to (\ref{eq:appendix_bound_4}), for all observed tasks $i=1,\dots,T$, we obtain that for any $\delta_i > 0$
\begin{equation}
\begin{footnotesize}  
    \begin{aligned}
        & \mathbb{P}_{\mathcal{T}_i\sim \mathcal{D}_i^m} \left\{\underset{z\sim \mathcal{D}_i}{\mathbb{E}} \underset{Q_{\rho, 0}\sim\mathcal{Q}}{\mathbb{E}} \underset{h\sim Q_{\rho, \Theta}}{\mathbb{E}} l(h,z) \leq \frac{1}{m_i}\sum_{j=1}^{m_i} \underset{Q_{\rho, 0}\sim\mathcal{Q}}{\mathbb{E}} \underset{h\sim Q_{\rho, \Theta}}{\mathbb{E}} l(h, z_{i,j}) \right.\\ 
        & \left. + \sqrt{\frac{1}{2(m_i-1)}\left(D(\mathcal{Q}\Vert\mathcal{P}) + \frac{1-\rho}{2}\Vert \Theta_i\Vert^2 + \log\frac{m_i}{\delta_i}\right)},\forall \mathcal{Q} \right\} \ge 1-\delta_i,
    \label{eq:appendix_bound_6}
    \end{aligned}
\end{footnotesize}
\end{equation}

\subsubsection{Second step} Similar to the first step, we use Theorem 2 with the following substitutions to bound the environment-level generalization error. Note that this is consistent with the previous work \cite{DBLP:conf/icml/AmitM18}, and we reformulated it here for completeness of proof. Let $(\mathcal{D}_i, m_i)$ be sampled from the task-distribution $\tau$ and $\mathcal{T}_i \sim D_i^{m_i}$, we denote iid samples as $(\mathcal{D}_i, m_i, \mathcal{T}_i), i=1,\dots,T$. The `hypotheses' are $f:=Q_{\rho, 0}$ and the `loss function' is $g(f, X):=\underset{h\sim Q_{\rho, \Theta}}{\mathbb{E}} \underset{z\sim \mathcal{D}}{\mathbb{E}}l(h,z)$. Let $\pi:=\mathcal{P}$ be prior distribution over hypothesis, the bound will hold uniformly for all distributions $\kappa:=\mathcal{Q}$, 
\begin{equation}
\begin{footnotesize}
    \begin{aligned}
        \mathbb{P}_{(\mathcal{D}_i, m_i)\sim \tau,\mathcal{T}_i\sim \mathcal{D}_i^{m_i},i=1,\dots,T} \left\{ \underset{(\mathcal{D}, m)\sim \tau}{\mathbb{E}} \underset{S\sim\mathcal{D}^m}{\mathbb{E}} \underset{Q_{\rho, 0}\sim \mathcal{Q}}{\mathbb{E}} \underset{h\sim Q_{\rho, \Theta}}{\mathbb{E}} \right. \\ 
        \left. \underset{z\sim\mathcal{D}}{\mathbb{E}}l(h,z) \leq \frac{1}{T}\sum_{i=1}^T \underset{Q_{\rho, 0}\sim\mathcal{Q}}{\mathbb{E}} \underset{h\sim Q_{\rho, \Theta}}{\mathbb{E}} \underset{z\sim\mathcal{D}_i}{\mathbb{E}}l(h, z) \right. \\
        \left. + \sqrt{\frac{1}{2(T-1)}\left(D(\mathcal{Q}\Vert\mathcal{P})+\log\frac{T}{\delta_0}\right)},\forall \mathcal{Q} \right\} \ge 1-\delta_0.
        \label{eq:appendix_bound_7}
    \end{aligned}
\end{footnotesize}
\end{equation}

Finally, denote the expected error of the meta-learner as $er(\mathcal{Q}, \tau) :=  \underset{(\mathcal{D}, m)\sim \tau}{\mathbb{E}} \underset{S\sim\mathcal{D}^m}{\mathbb{E}} \underset{Q_{\rho, 0}\sim \mathcal{Q}}{\mathbb{E}} \underset{h\sim Q_{\rho, \Theta}}{\mathbb{E}} \underset{z\sim\mathcal{D}}{\mathbb{E}}l(h,z)$ and empirical error of each task as $\hat{er}(\mathcal{Q}, \mathcal{T}) := \underset{h\sim Q_{\rho, \Theta}}{\mathbb{E}} \underset{z\sim\mathcal{D}_i}{\mathbb{E}}l(h, z)$ respectively. We will bound the probability of the event that is the intersection of the events in (\ref{eq:appendix_bound_6}) and (\ref{eq:appendix_bound_7}) by using the union bound. For any $\delta > 0$, set $\delta_0:=\frac{\delta}{2}$ and $\delta_i:=\frac{\delta}{2T}$ for $i=1,\dots,T$, the following hold.
\begin{equation}
\begin{footnotesize}
    \begin{aligned}
    er(\mathcal{Q}) \leq \frac{1}{T}\sum_{i=1}^{T} \underset{Q_{\rho, 0} \sim \mathcal{Q}}{\mathbb{E}} \hat{er}_i(Q_{\rho,\Theta_i}, \mathcal{T}_i) + \sqrt{\frac{D(\mathcal{Q}\Vert \mathcal{P}) + \log \frac{2T}{\delta}}{2(T-1)}}\\ 
    + \frac{1}{T}\sum_{i=1}^{T}\sqrt{\frac{D(\mathcal{Q}\Vert \mathcal{P}) + \log \frac{2Tm_i}{\delta} + \frac{1-\rho}{2}\Vert \Theta_i \Vert^2}{2(m_i-1)}}, 
    \end{aligned}
\end{footnotesize}
\end{equation}
which completes the inductive proof. 

\section*{Acknowledgements}
This research was supported in part by the Natural Science Foundation of China (No. 62106129, 62176139, and 62177031), the Natural Science Foundation of Shandong Province (No. ZR2021QF053, ZR2021ZD15), and the China Postdoctoral Science Foundation (No. 2021TQ0195, 2021M701984).


\bibliographystyle{IEEEtran}
\bibliography{refs}

\end{document}